\pdfoutput=1

\documentclass[11pt, oneside]{article}    

\newcommand{\minenet}{MineRL}

\usepackage{geometry}
\usepackage{graphicx}
\usepackage{color}
\usepackage[numbers]{natbib}
\usepackage{caption}
\usepackage{xcolor}
\usepackage{booktabs}
\usepackage{wrapfig}
\usepackage{amssymb}
\usepackage{subcaption}
\newcommand{\footremember}[2]{%
\footnote{#2}
\newcounter{#1}
\setcounter{#1}{\value{footnote}}%
}
\newcommand{\footrecall}[1]{%
\footnotemark[\value{#1}]%
}

\usepackage{todonotes}
\usepackage{hyperref}
\usepackage{fancyhdr}

\title{NeurIPS 2020 Competition: The MineRL Competition on Sample Efficient Reinforcement Learning using Human Priors }

\author{William H. Guss\footremember{lead}{Lead organizer: \texttt{wguss@cs.cmu.edu}}\footremember{cmu}{Affiliation: Carnegie Mellon University}\footremember{openai}{Affiliation: OpenAI Inc.}
\and Mario Ynocente Castro\footremember{eq}{
    \textbf{Equal contribution: Organizer names are ordered alphabetically}, with the exception of the lead organizer. Competitions are extremely complicated endeavors involving a huge amount of organizational overhead from the development of complicated software packages to event logistics and evaluation. It is impossible to estimate the total contributions of all involved at the onset.
}\footremember{pfn}{Affiliation: Preferred Networks, Inc.}
\and Sam Devlin\footrecall{eq} \footremember{ms}{Affiliation: Microsoft Research}
\and Brandon Houghton\footrecall{eq} \footrecall{openai}
\and Noboru Sean Kuno\footrecall{eq} \footrecall{ms}
\and Crissman Loomis\footrecall{eq} \footrecall{pfn}
\and Stephanie Milani\footrecall{eq} \footrecall{cmu}
\and Sharada Mohanty\footrecall{eq} \footremember{ai}{Affiliation: AIcrowd SA}
\and Keisuke Nakata\footrecall{eq} \footrecall{pfn}
\and Ruslan Salakhutdinov\footrecall{eq} \footrecall{cmu}
\and John Schulman\footrecall{eq} \footrecall{openai}
\and Shinya Shiroshita\footrecall{eq} \footrecall{pfn}
\and Nicholay Topin\footrecall{eq} \footrecall{cmu}
\and Avinash Ummadisingu\footrecall{eq} \footrecall{pfn}
\and Oriol Vinyals\footrecall{eq} \footremember{dm}{Affiliation: DeepMind}
}

\date{}

\begin{document}
\maketitle
\vspace{-20pt}

\section*{Competition Overview}
    Although deep reinforcement learning has led to breakthroughs in many difficult domains, these successes have required an ever-increasing number of samples. As state-of-the-art reinforcement learning (RL) systems require an ever-increasing number of samples, their development is restricted to a continually shrinking segment of the AI community. Likewise, many of these systems cannot be applied to real-world problems, where environment samples are expensive. Resolution of these limitations requires new, sample-efficient methods. To facilitate research in this direction, we propose the \emph{MineRL 2020 Competition on Sample Efficient Reinforcement Learning using Human Priors}\footnote{\url{https://www.aicrowd.com/challenges/neurips-2020-minerl-competition}}.

    The primary goal of the competition is to 
        foster the development of algorithms which can efficiently leverage human demonstrations to drastically reduce the number of samples needed to solve complex, hierarchical, and sparse environments. 
        To that end, participants will compete under a limited environment sample-complexity budget to develop systems which solve the MineRL \texttt{ObtainDiamond} task, a sequential decision making environment requiring long-term planning, hierarchical control, and efficient exploration methods.  
    Participants will be provided the \emph{MineRL-v0} dataset~\cite{gussminerlijcai2019}, a large-scale collection of over 60 million state-action pairs of human demonstrations that can be resimulated into embodied agent trajectories with arbitrary modifications to game state and visuals.

    The competition is structured into two rounds in which competitors 
        are provided several paired versions of the dataset and environment with different game textures and shaders.
    At the end of each round, competitors will submit containerized 
        versions of their learning algorithms to the AIcrowd platform where they will then be trained from scratch on a hold-out  dataset-environment pair for a total of 4-days on a pre-specified hardware platform. 
        Each submission will then be automatically ranked according to the final performance of the trained agent.

    This challenge is a follow-up to our NeurIPS 2019 MineRL competition~\cite{gussminerlneurips2019}, which yielded over 1000 registered participants and over 662 full submissions. The competition benchmark, RL environment, and dataset framework were downloaded over 52,000 times in 26+ countries~\cite{milani2020minerl}. 
        In this iteration, we will implement new features to expand the scale and reach of the competition. In response to the feedback of the previous participants, we are introducing a second minor track focusing on solutions \textit{without access to environment interactions} of any kind except during test-time. Both tracks will follow the same two-round schedule.
        Last year's top submissions developed novel methods advancing inverse reinforcement learning, hierarchical imitation learning, and more. In the forthcoming competition, we anticipate an even larger research impact. With the addition of action-space randomization and desemantization of observations and actions, we believe that the most successful competition submissions will be highly task and domain agnostic.

\subsection*{Keywords}
Reinforcement Learning, Imitation Learning, Sample Efficiency, Games, MineRL, Minecraft.
\subsection*{Competition Type} Regular.

\newpage 

\section{Competition Description}

\subsection{Background and Impact}

Many of the recent, most celebrated successes of artificial intelligence (AI), such as AlphaStar~\cite{starcraft2019}, AlphaGo~\cite{silver2017mastering}, OpenAI Five~\cite{berner2019dota}, and their derivative systems~\cite{alphazero}, utilize deep reinforcement learning to achieve human or super-human level performance in sequential decision-making tasks.
    These improvements to the state-of-the-art have thus far required exponentially increasing computational power to achieve such performance~\cite{amodei_hednandez_2018}.
        In part, this is due to an increase in the computation required per environment-sample; however, the most significant change is the number of environment-samples required for training. 
            For example, DQN~\cite{mnih2015human}, A3C~\cite{mnih2016asynchronous}, and Rainbow DQN~\cite{hessel2018rainbow} have been applied to ATARI 2600 games~\cite{bellemare2013arcade} and require from 44 to over 200 million frames (200 to over 900 hours) to achieve human-level performance. 
            On more complex domains: OpenAI Five utilizes 11,000+ years of Dota 2 gameplay~\cite{openai_2018}, AlphaGoZero uses 4.9 million games of self-play in Go~\cite{silver2017mastering}, and AlphaStar uses 200 years of StarCraft~II gameplay~\cite{deepmind}. 
    Due to the growing computational requirements, a shrinking portion of the AI community has the resources to improve these systems and reproduce state-of-the-art results. 
        Additionally, the application of many reinforcement learning techniques to real-world challenges, such as self-driving vehicles, is hindered by the raw number of required samples.
        In these real-world domains, policy roll-outs can be costly and simulators are not yet accurate enough to yield policies robust to real-world conditions.

One well-known way to reduce the environment sample-complexity of the aforementioned methods is to leverage human priors and demonstrations of the desired behavior. 
    Techniques utilizing trajectory examples, such as imitation learning and Bayesian reinforcement learning, have been successfully applied to older benchmarks and real-world problems where samples from the environment are costly.  
        In many simple games with singular tasks, such as the Atari 2600~\cite{bellemare2013arcade}, OpenAI Gym~\cite{gym}, and TORCS environments\footnote{\url{https://github.com/ugo-nama-kun/gym_torcs}}, imitation learning can drastically reduce the number of environment samples needed through pretraining and hybrid RL techniques~\cite{cruz2017pre,gao2018reinforcement,hester2018deep,panse2018imitation}.
        Further, in some real-world tasks, such as robotic manipulation~\cite{finn2016guided,finn2017one} and self-driving~\cite{bojarski2016end}, in which it is expensive to gather a large number of samples from the environment, imitation-based methods are often the only means of generating solutions using few samples.
    Despite their success, these techniques are still not sufficiently sample-efficient for application to many real-world domains.

\begin{figure}
    \centering
    \includegraphics[width=0.8\textwidth]{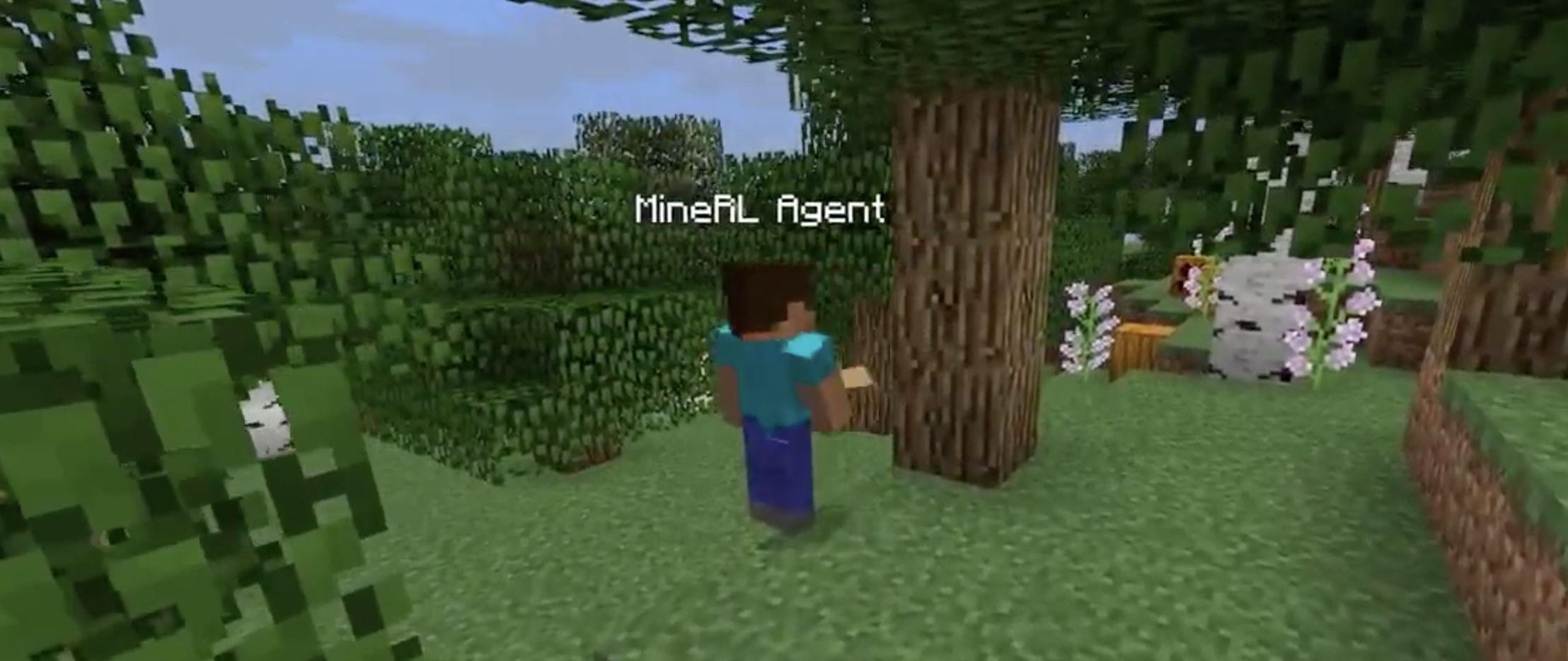}
    \caption{\small{The top agent from the MineRL 2019 competition mining the first item required to eventually obtain a diamond.}}
\end{figure}

\paragraph{Impact.} To that end, the central aim of our proposed competition is the advancement and development of novel, sample-efficient methods which leverage human priors for sequential decision-making problems. 
    Due to the competition's design, organizational team, and support, we are confident that the competition will catalyze research towards the deployment of reinforcement learning in the real world, democratized access to AI/ML, and reproducibility. 
    By enforcing constraints on the computation and sample budgets of the considered techniques, we believe that the methods developed during the competition will broaden participation in deep RL research by lowering the computational barrier to entry. 

While computational resources inherently have a cost barrier, large-scale, open-access datasets can be widely used. 
    To that end, we center our proposed competition around techniques which leverage the \minenet{} dataset~\cite{gussminerlijcai2019}.
    To maximize the development of domain-agnostic techniques that enable the application of deep reinforcement learning to sample-limited, real-world domains, such as robotics, we carefully developed a novel data-pipeline and hold-out environment evaluation scheme with AIcrowd to prevent the over-engineering of submissions to the competition task.

Crucially, the competition will stimulate a broad set of new techniques in reinforcement and imitation learning. 
    In the previous NeurIPS 2019 iteration of the competition, competitors developed several new algorithms and approaches to tackle the challenge in spite of the difficult sample-complexity limitations~\cite{gussminerlneurips2019}.
        Ranging from hierarchical imitation methods to novel inverse reinforcement learning techniques, the research impact of the competition was broad in scope, yielding a diverse set of solutions~\cite{milani2020minerl}. 
    With the addition of new competition features and refined submission and evaluation pipelines (see Section~\ref{sec:novelty}),
    we anticipate this year's competition to garner further research progress of relevance to the NeurIPS community.

Our competition will further attract a large number of participants from within and outside of the NeurIPS community. 
    Given the broad interest and participation in the previous year (attracting over 1000 registered participants with a total of 662 full submissions\footnote{\url{https://www.aicrowd.com/challenges/neurips-2019-minerl-competition}}), our extensive media coverage~\cite{hsu_2019,shead_2019,synced_2019,vincent_2019}, and improvements to user-experience, we expect the number of participants to grow to 1300 users and the number of successful submission to increase to over 1000 agents. To effectuate this growth, we will deliver several improvements over prior years. First, we plan to drastically simplify the submission process and provide thorough multi-media documentation to increase the conversion-rate from registration to submission. Further, we intend on providing more compelling visualizations for the competitors' submissions, generating external interest from outside of the research community. Expanding on media coverage and outreach channels from last year, we will utilize mailing lists and social media announcements to retain the previous competitor pool and expand our user-base to new demographics.
        Moreover, the expansion of our competition to multiple tracks supporting pure imitation learning and hybridized imitation and reinforcement learning submissions will broaden the the appeal of our competition as a vehicle for researching and developing new methods.

The proposed competition is ambitious, so we have taken meaningful steps to ensure its smooth execution. 
Specifically, we are currently securing several crucial partnerships with organizations and individuals. 
During the MineRL 2019 competition, our primary partner, Microsoft Research, provided significant computational resources to enable direct, fair evaluation of the participants' training procedures. 
We developed a relationship with AIcrowd to provide the submission orchestration platform for our competition, as well as continued support throughout the competition to ensure that participants can easily submit their algorithms. Additionally, we partnered with Preferred Networks in the previous iteration of this competition to provide a set of standard baseline implementations, which include many state of the art reinforcement learning and imitation learning techniques. By leveraging our previous partnerships and developing new ones, we expect to largely increase the scale, success, and impact of the competition.


\subsubsection{Domain Interest}

\begin{figure}
    \begin{center}
        \includegraphics[width=0.9\textwidth]{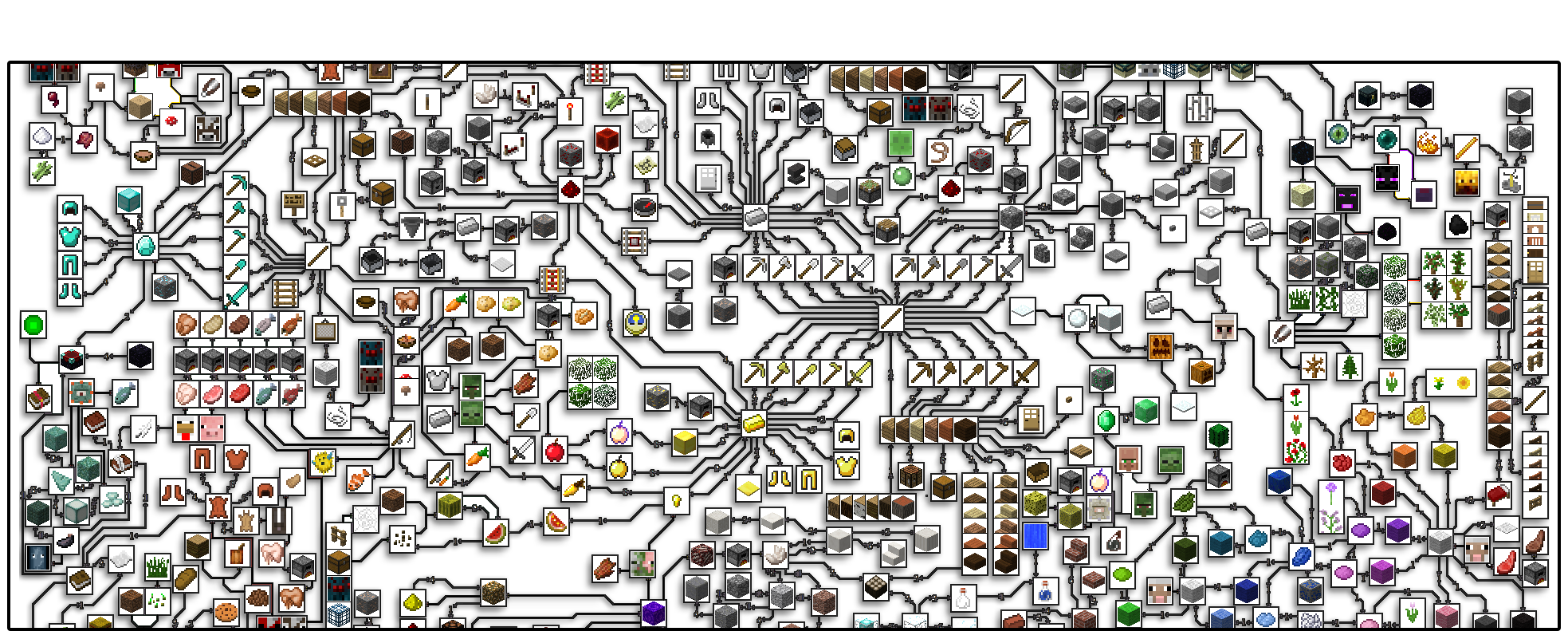}
        \caption{\small{A subset of the Minecraft item hierarchy (totaling 371
        unique items). Each node is a unique Minecraft item, block, or non-player character, and a directed edge between two nodes denotes that
        one is a prerequisite for another. Each item presents is own unique
        set of challenges, so coverage of the full hierarchy by one player
        takes several hundred hours.}}
        \label{fig:hierarchicality}
    \end{center}
    \vspace{-19pt}
\end{figure}

Minecraft is a compelling domain for the development of reinforcement and imitation learning methods because of the unique challenges it presents: Minecraft is a 3D, first-person, open-world game centered around the gathering of resources and creation of structures and items. 
Notably, the procedurally-generated world is composed of discrete blocks that allow modification; over the course of gameplay, players change their surroundings by gathering resources (such as wood from trees) and constructing structures (such as shelter and storage).
Since Minecraft is an embodied domain and the agent's surroundings are varied and dynamic, it presents many of the same challenges as real-world robotics domains. 
Therefore, solutions created for this competition are a step toward applying these same methods to real-world problems.

Furthermore, there is existing research interest in Minecraft. 
With the development of Malmo~\cite{johnson2016malmo}, a simulator for Minecraft, the environment has garnered great research interest:
many researchers~\cite{oh2016control,shu2017hierarchical,tessler2017deep} have leveraged Minecraft's massive hierarchality and expressive power as a simulator to make great strides in language-grounded, interpretable multi-task option-extraction, hierarchical lifelong learning, and active perception. 
However, much of the existing research utilizes toy tasks in Minecraft, often restricted to 2D movement, discrete positions, or artificially confined maps unrepresentative of the intrinsic complexity that human players typically face. 
These restrictions reflect the difficulty of the domain, the challenge of coping with fully-embodied human state- and action-spaces, and the complexity exhibited in optimal human policies. 

Our competition and the utilization of the large-scale \minenet-v0 dataset of human demonstrations will serve to catalyze research on this domain in two ways: (1) our preliminary results indicate that through imitation learning, basic reinforcement learning approaches can finally deal directly with the full, unrestricted state- and action-space of Minecraft;  and (2) due to the difficult and crucial research challenges exhibited on the primary competition task, \texttt{ObtainDiamond}, we believe that the competition will bring work on the Minecraft domain to the fore of sample-efficient reinforcement learning research.

\subsection{Novelty} \label{sec:novelty}

This year's MineRL Competition is a follow-up to the first MineRL competition held at NeurIPS 2019.
We continue to encourage the development of general learning algorithms which must perform well within a \emph{strict} computation and environment-sample budget. 
Based on community feedback and our retrospection, we are making the following improvements to this year's competition:

\begin{itemize}
    \item To further encourage competitors to develop generalizable methods, we are updating the rules on manually specified policies and pre-processing of the action space. In particular, we are improving the clarity of the rules, and we are no longer allowing submissions to manually specify action choices. In the previous MineRL Competition, actions could be specified by the competitors as long as the setting did not depend on an aspect of the state.
    \item To ensure that competitors do not exploit the semantic meanings attached to the action or observation labels, we embed both the action and non-POV observations individually into latent spaces using auto-encoders. This makes it difficult to manually specify meaningful actions, or hard-code behaviors based on observations by providing obfuscated vectors for the action and observation spaces. The networks trained to embed and recover actions and observations ensure that the original actions and observations are recoverable from the embedded space, but also that entire embedded domain maps onto the original space. Additionally, this embedding is changed in subsequent rounds to ensure generalizability. Previously, labels were modified during evaluation, but they still carried semantic meaning and were not fully obfuscated.
    \item To further encourage the use of methods that learn from demonstrations, we are adding a second track to the competition. This track will follow the same restrictions as the original track, but competitors will not be permitted to use the environment during training. By adding this track, competitors interested in learning from demonstrations can compete without being disadvantaged compared to those who also use reinforcement learning. Additionally, this track will help quantify the performance attainable using only demonstrations.
\end{itemize}

Our competition focuses on the application of reinforcement learning and imitation learning to a domain in Minecraft. 
As a result, it is related to competitions which focus on these three aspects.
We briefly identify related competitions and describe the key differences between our proposed competition and the other competitions.

\paragraph{Reinforcement Learning.} Prior to our competition series, reinforcement learning competitions have focused on the development of policies or meta-policies that perform well on complex domains or generalize across a distribution of tasks~\cite{kidzinski2018learning, nichol2018gotta, perez2019multi}. 
However, the winning submissions of these competitions are often the result of massive amounts of computational resources or highly specific, hand-engineered features. 
In contrast, our competition directly considers the efficiency of the training procedures of learning algorithms.

We evaluate submissions solely on their ability to perform well within a \emph{strict} computation and environment-sample budget. 
Moreover, we are uniquely positioned to propose such a competition due to the nature of our human demonstration dataset and environment: our dataset is constructed by directly recording the game-state as human experts play, so we are able to later make multiple renders of both the environment and data with varied lighting, geometry, textures, and game-state dynamics, thus yielding development, validation, and hold-out evaluation dataset/environment pairs. 
As a result, competitors are naturally prohibited from hand-engineering or warm-starting their learning algorithms and winning solely due to resource advantages. 

\paragraph{Imitation Learning.} To our knowledge, no competitions have explicitly focused on the use of imitation learning alongside reinforcement learning. 
This is in large part due to a lack of large-scale, publicly available datasets of human or expert demonstrations. 
Our competition is the first to explicitly involve and encourage the use of imitation learning to solve the given task, and in that capacity, we release the largest-ever dataset of human demonstrations on an embodied domain.
The large number of trajectories and rich demonstration-performance annotations enable the application of many standard imitation learning techniques and encourage further development of new ones that use hierarchical labels, varying agent performance levels, and auxiliary state information.

\paragraph{Minecraft.} 
A few competitions have previously used Minecraft due to its expressive power as a domain. 
The first one was The Malm\"{o} Collaborative AI Challenge\footnote{\url{https://www.microsoft.com/en-us/research/academic-program/collaborative-ai-challenge}}, in which agents worked in pairs to solve a collaborative task in a decentralized manner. 
Later, C. Salge et al.~\cite{salge2018generative} organized the Generative Design in Minecraft (GDMC): Settlement Generation Competition, in which participants were asked to implement methods that would procedurally build complete cities in any given, unknown landscape. 
These two contests highlight the versatility of this framework as a benchmark for different AI tasks.

In 2018, Perez-Liebana et al.~\cite{perez2019multi} organized the Multi-Agent Reinforcement Learning in Malm\"{O} (MARL\"{O}) competition. 
This competition pitted groups of agents to compete against each other in three different games.
Each of the games was parameterizable to prevent the agents from overfitting to specific visuals and layouts. 
The objective of the competition was to build an agent that would learn, in a cooperative or competitive multi-agent task, to play the games in the presence of other agents. 
The MARL\"{O} competition successfully attracted a large number of entries from both existing research institutions and the general public, indicating a broad level of accessibility and excitement for the Minecraft domain within and outside of the existing research community.

In comparison with previous contests, the MineRL series of competitions tackles one main task and provides a massive number of hierarchical subtasks and demonstrations (see Section~\ref{sec:data}). 
The main task and its subtasks are not trivial; however, agent progress can be easily measured, which allows for a clear comparison between submitted methods. 
Further, the target of the competition series is to promote research on \textit{efficient learning}, focusing directly on the sample- and computational-efficiency of the submitted algorithms~\cite{houghton2020guaranteeing}.

\subsection{Data}\label{sec:data}

For this competition, we utilize two main components: a set of sequential decision making environments in Minecraft and a corresponding public large-scale dataset of human demonstrations. Through an online server which replicates these environments, we continue to engage the Minecraft community to add additional demonstrations to this dataset.

\subsubsection{Environment}

We define \emph{one primary competition environment}, \texttt{ObtainDiamond}, and six other auxiliary environments that encompass a significant portion of human Minecraft play. 
We select these environment domains to highlight many of the hardest challenges in reinforcement learning, such as sparse rewards, long reward horizons, and efficient hierarchical planning.

\paragraph{Primary Environment.}
As with last year's competition, the main task of this year's competition is solving the \texttt{Obtain} \texttt{Diamond} environment. 
In this environment, the agent begins in a random starting location without any items, and is tasked with obtaining a diamond. 
The agent receives a high reward for obtaining a diamond and smaller, auxiliary rewards for obtaining prerequisite items. 
Episodes end due to the agent dying, successfully obtaining a diamond, or reaching the maximum step count of 18000 frames (15 minutes).

\paragraph{Auxiliary Environments.}
\begin{figure}
    \begin{centering}
        \vspace{-10pt}
        \includegraphics[width=0.49\textwidth]{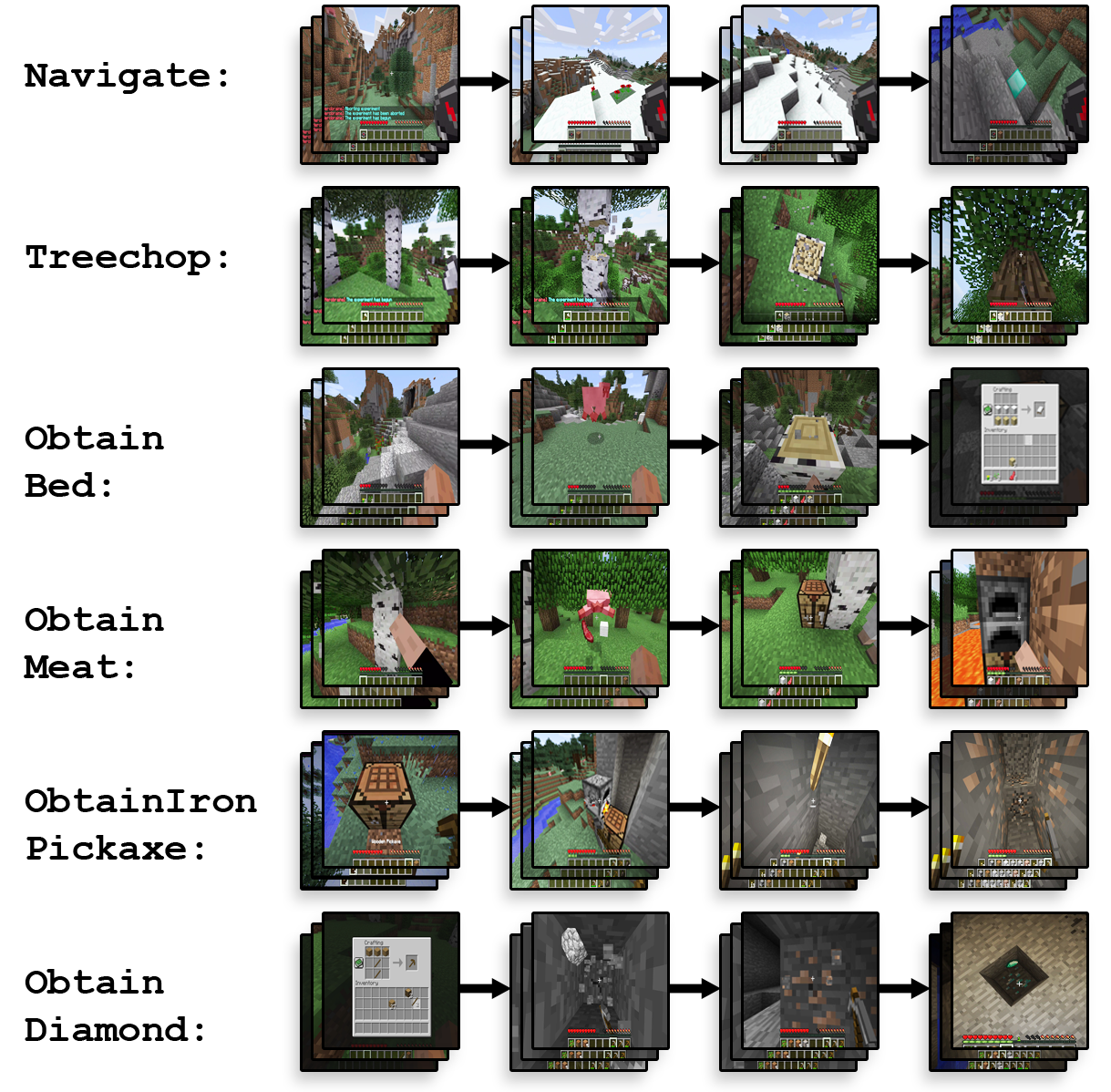} 
        \caption{\small Images of various stages of six of seven total environments.}
        \label{fig:tasks}
        \vspace{-10pt}
    \end{centering}
\end{figure}

The \texttt{ObtainDiamond} environment is a difficult environment; diamonds only exist in a small portion of the world and are 2-10 times rarer than other ores in Minecraft.
Furthermore, obtaining a diamond requires many prerequisite items. 
It is practically impossible for an agent to obtain a diamond via naive random exploration.

We provide six auxiliary environments (in four families), which we believe will be useful for solving \texttt{ObtainDiamond}:
\begin{enumerate}
    \item \texttt{Navigate}: In this environment, the agent must move to a goal location, which represents a basic primitive used in many tasks in Minecraft. 
    In addition to standard observations, the agent has access to a ``compass'' observation,
    which points to a set location, 64 meters from the start location. 
    The agent is given a sparse reward (+100 upon reaching the goal, at which point the episode terminates). 
    We also support a dense, reward-shaped version of Navigate, in which the agent receives reward every tick corresponding to the change in distance between the agent and the goal.

    \item \texttt{Treechop}: In this environment, the agent must collect wood, a key resource in Minecraft and the first prerequisite item for diamonds.
    The agent begins in a forest biome (near many trees) with an iron axe for cutting trees. The agent is given +1 reward for obtaining each unit of wood, and the episode terminates once the agent obtains 64 units or the step limit is reached.
    
    \item \texttt{Obtain<Item>}:
    We include three additional obtain environments, similar to that of \texttt{ObtainDiamond},
    but with different goal items to obtain. They are:
    
\begin{enumerate}
    \item \texttt{CookedMeat}: cooked meat of a (cow, chicken, sheep, or pig), which is necessary for survival in Minecraft. In this environment, the agent is given a specific kind of meat to obtain.
    \item \texttt{Bed}: made out of dye, wool, and wood, an item that is also vital to Minecraft survival. In this environment, the agent is given a specific color of bed to create.
    \item \texttt{IronPickaxe}: is a final prerequisite item in obtaining a diamond.
    It is significantly easier to solve than \texttt{ObtainDiamond}: iron is
    20 times more common in the Minecraft world than diamonds, and this environment is typically solved by
    humans in less than 10 minutes.
\end{enumerate}

    \item \texttt{Survival}: This environment is the standard, open-ended game mode used by most human players when playing the game casually.
    There is no specified reward function, but data from this environment can be used
    to help train agents in more structured tasks, such as \texttt{ObtainDiamond}.

\end{enumerate}

\subsubsection{Dataset}

 \begin{figure}
    \begin{center}
        \includegraphics[width=0.47\textwidth]{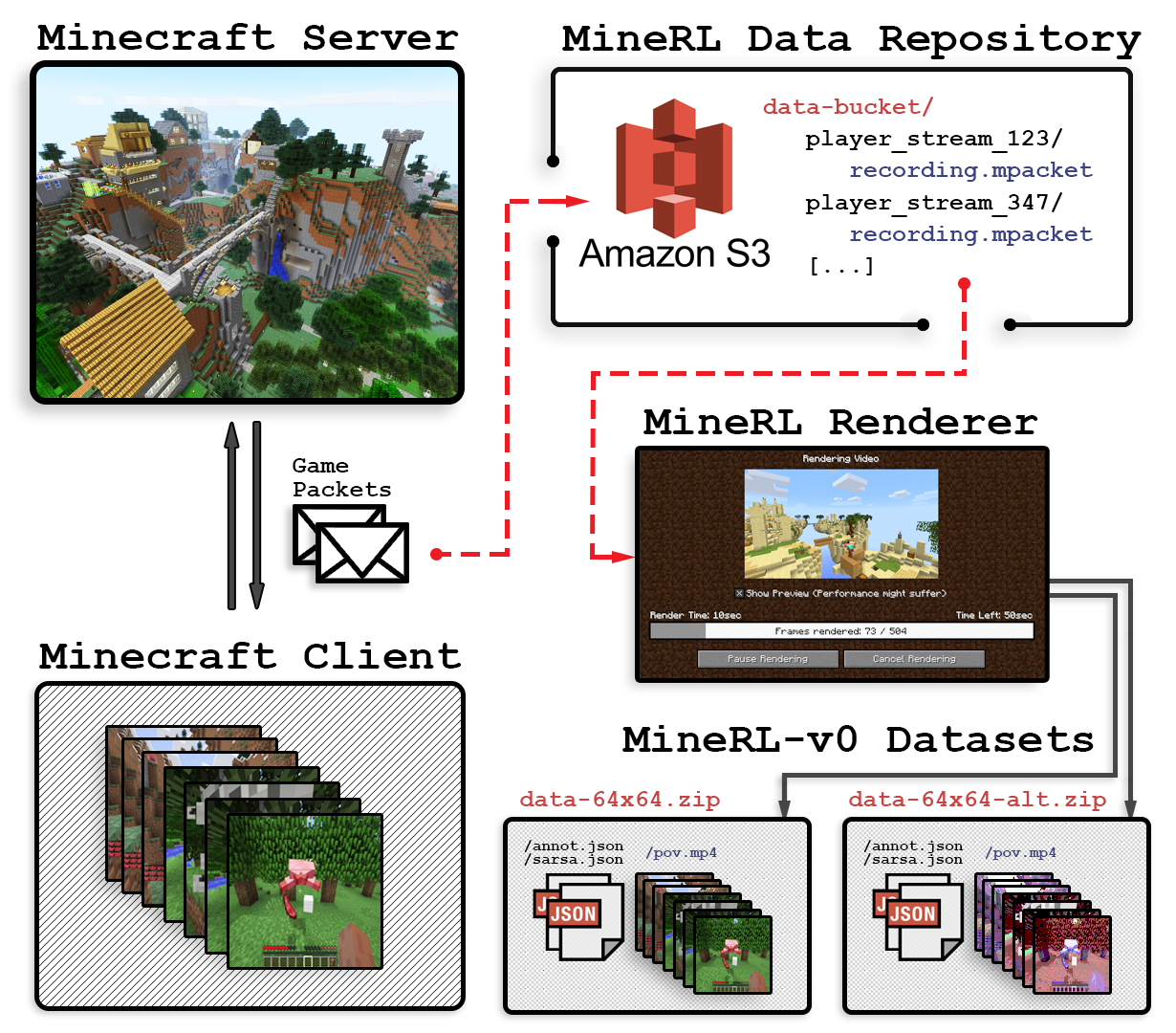}
         \caption{\small A diagram of the MineRL data collection platform. Our
        system renders demonstrations from packet-level data, so we can
        easily rerender our data with different parameters.}
        \label{fig:platform_diagram}
    \end{center}
\end{figure}

The \minenet-v0 dataset consists of over 60 million state-action-(reward) tuples of recorded human demonstrations
over the seven environments mentioned above~\cite{gussminerlijcai2019}. In addition, we are actively working with the community to record additional human demonstrations. 
Trajectories are contiguously sampled every Minecraft game tick (at 20 game ticks per second). Each state is comprised of an RGB video frame of the player's point-of-view and a comprehensive set of features from the game-state at that tick:
player inventory, item collection events, distances to objectives, player attributes (health,
level, achievements), and details about the current GUI the player has open. The action
recorded at each tick consists of: all the keyboard presses, the change in view pitch and yaw
(mouse movements), player GUI interactions, and agglomerative
actions such as item crafting.

Accompanying the human trajectories are a large set of automatically generated annotations.
For all of the environments, we include metrics which indicate the quality of the demonstration, such as timestamped rewards, number of no-ops, number of deaths, and total score.
Additionally, trajectory meta-data includes timestamped markers for hierarchical labelings; e.g. when a house-like structure is built or certain objectives such as chopping down a tree are met. Data is made available both in the competition materials as well as through a standalone website\footnote{\url{http://minerl.io}}.

\subsubsection{Data Collection}
    In the previous MineRL competition, we used our novel platform for the collection of player trajectories in Minecraft, enabling the construction of the \minenet-v0 dataset.  In this second iteration of the competition, we will continue to utilize the platform with the hope of drastically expanding the existing dataset.
    As shown in Figure~\ref{fig:platform_diagram}, our platform consists of 
    (1) \emph{a public game server and website}, where we obtain permission to record trajectories of Minecraft players in natural gameplay;
    (2) \emph{a custom Minecraft client plugin}, which records all packet level communication between the client and the server, so we can re-simulate and re-render human demonstrations with modifications to the game state and graphics; 
    and (3) \emph{a data processing pipeline}, which enables us to produce automatically annotated datasets of task demonstrations.

    \paragraph{Data Acquisition.}
    Minecraft players find the \minenet~server on standard Minecraft server lists. 
        Players first use our webpage to provide IRB\footnote{The data collection study was approved by Carnegie Mellon University's institutional review board as \texttt{STUDY2018\_00000364}.} consent to have their gameplay anonymously recorded. Then, they download a plugin for their Minecraft client, which records and streams users' client-server game packets to the \minenet~data repository.
    When playing on our server, users select an environment to solve and receive in-game currency proportional to the amount of reward obtained. 
    For the \texttt{Survival} environment (where there is no known reward function), players
receive rewards only for duration of gameplay, so as not to impose an artificial reward function.

\paragraph{Data Pipeline.}
Our data pipeline allows us to resimulate recorded trajectories into several algorithmically consumable formats. 
    The pipeline serves as an extension to the core Minecraft game code and synchronously sends each recorded packet from the \minenet{} data repository to a Minecraft client using our custom API for automatic annotation and game-state modification. 
    This API allows us to add annotations based on any aspect of the game state accessible from existing Minecraft simulators. 
    Notably, it allows us to rerender the same data with different textures, shaders, and lighting-conditions which we use to create test and validation environment-dataset pairs for this competition.
 
\subsubsection{Data Usefulness}\label{sec:data_use}

\begin{wrapfigure}{r}{0.5\textwidth}

    \vspace{-25pt}
    \begin{center}
        \includegraphics[width=0.49\textwidth]{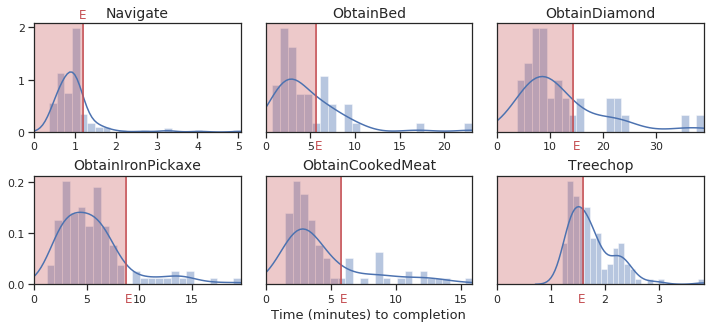}
    \caption{\small Normalized histograms of the lengths of human demonstration on various \minenet~tasks. The red {\color{red} \tiny\sf{E}} denotes the upper threshold  for expert play on each task.}
    \label{fig:human_quality}
    \end{center}
    \vspace{-20pt}
\end{wrapfigure}

\paragraph{Human Performance.}

    A majority of the human demonstrations in the dataset fall within the range of expert level play.
    Figure \ref{fig:human_quality} shows the distribution over trajectory length for each environment.
    The red region in each histogram denotes the range of times which correspond to play at an expert level, computed as the average time required for task completion by players with at least five years of Minecraft experience. 
    The large number of expert samples and rich labelings of demonstration 
    performance enable application of many standard imitation learning techniques which assume optimality of the base policy. 
    In addition, beginner and intermediate level trajectories 
    allow for the further development of techniques that leverage imperfect demonstrations.
    
    \begin{wrapfigure}{r}{0.5\textwidth}
        \begin{center}
            \vspace{-15pt}
            \includegraphics[width=0.47\textwidth]{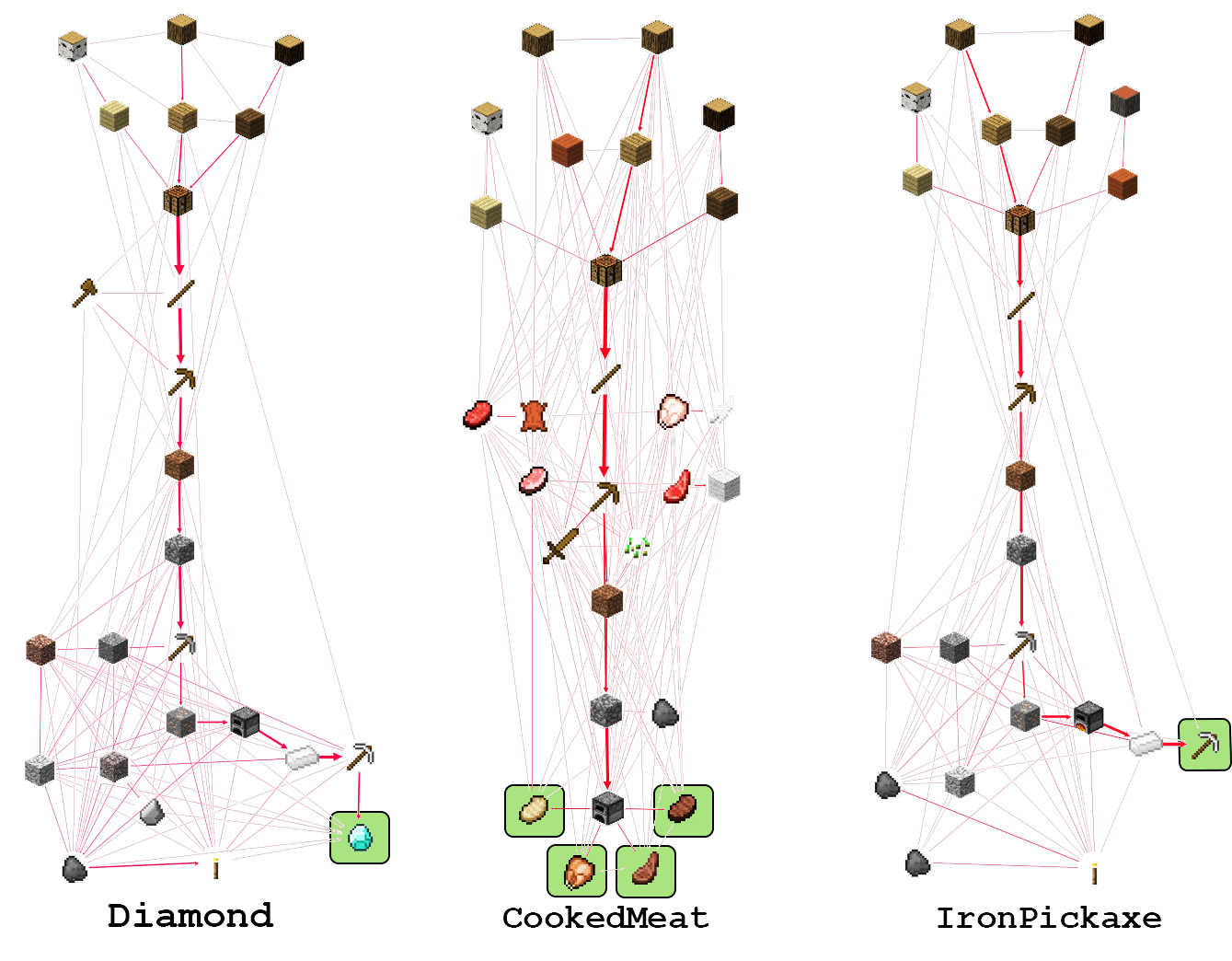}
            \caption{\small Item precedence frequency graphs for \texttt{ObtainDiamond} (left), \texttt{ObtainCookedMeat} (middle), and \texttt{ObtainIronPickaxe} (right). The thickness of each line indicates the number of times a player collected item $A$ then subsequently item $B$. } \label{fig:task_hist}
            \vspace{-18pt}
        \end{center}
    \end{wrapfigure}

\paragraph{Hierarchality.} 

As shown in Figure~\ref{fig:hierarchicality}, Minecraft is deeply hierarchical, and the \minenet~data collection platform is designed to capture these hierarchies both explicitly and implicitly. 
Due to the subtask labelings provided in \minenet-v0, we can inspect and quantify the extent to which these environments overlap. 
Figure~\ref{fig:task_hist} shows precedence frequency graphs constructed from \minenet{} trajectories on the \texttt{ObtainDiamond}, \texttt{Obtain}\texttt{CookedMeat}, and \texttt{ObtainIronPickaxe} tasks.
In order to complete the \texttt{ObtainDiamond} task, an agent must complete the sub-goals of obtaining wood and stone, as well as constructing crafting tables and furnaces.
These subtasks also appear in \texttt{ObtainIronPickaxe} and \\ \texttt{ObtainCookedMeat}.
There is even greater overlap between \texttt{ObtainDiamond}
and \texttt{ObtainIronPickaxe}: most of the item hierarchy for \texttt{ObtainDiamond} consists of the hierarchy for \texttt{ObtainIronPickaxe}.

\begin{figure}
    \centering
    \begin{subfigure}[b]{0.87\textwidth}
    \centering 
        \includegraphics[width=0.87\textwidth]{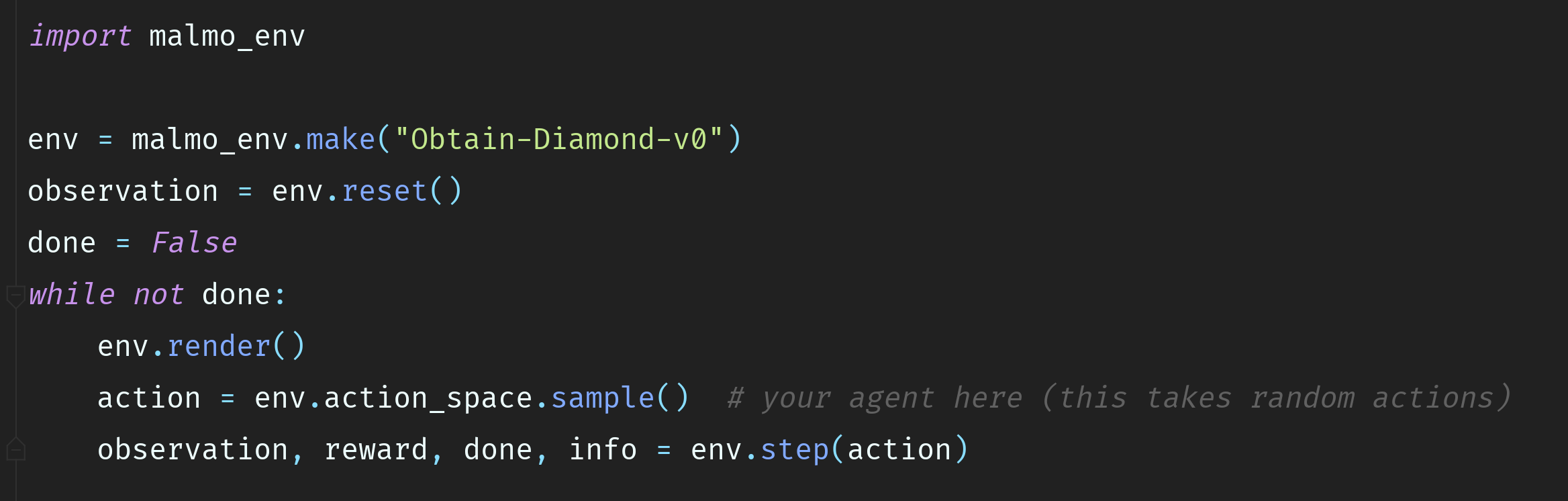}
        \caption{\small Running a single episode of a random agent in \texttt{ObtainDiamond}. } 
        \label{fig:env_code}
    \end{subfigure}

    \begin{subfigure}[b]{0.87\textwidth}
    \centering
        \includegraphics[width=0.87\textwidth]{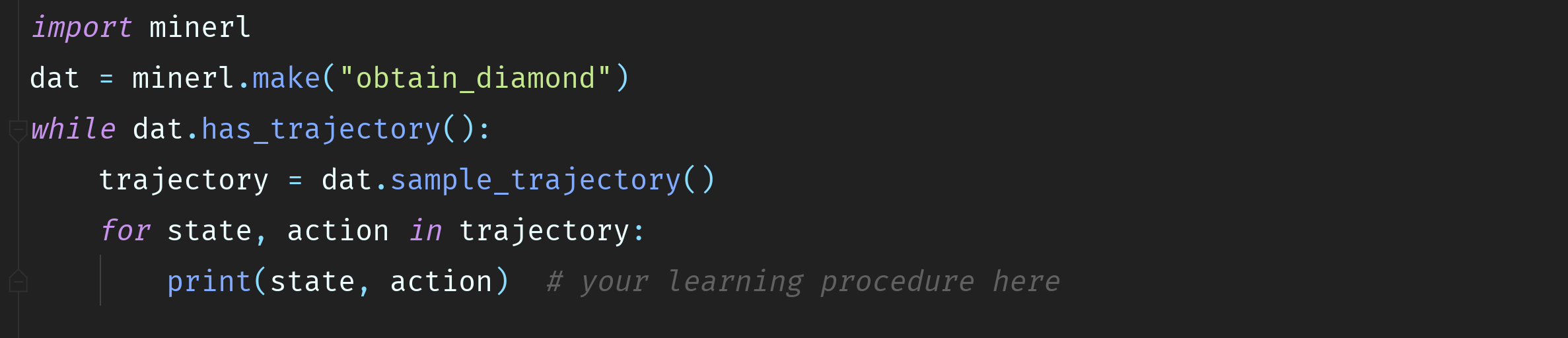}
        \caption{\small Utilizing individual trajectories of the \minenet dataset.} 
        \label{fig:mineral_code_1}
    \end{subfigure}

\begin{subfigure}[b]{0.87\textwidth}
    \centering
        \includegraphics[width=0.87\textwidth]{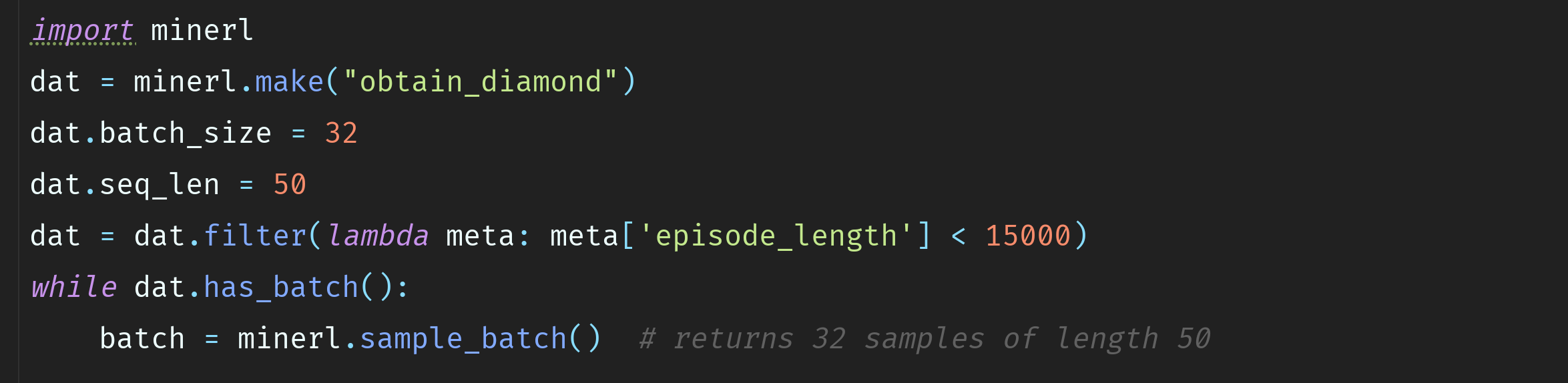}
        \caption{\small Using the \minenet  wrapper to filter demonstrations based on metadata} 
        \label{fig:mineral_code_2}
\end{subfigure}
\caption{\small Example code showing how to interact with MineRL data and environment.}
\label{fig:example_code}
\end{figure}

\paragraph{Interface}
Interacting with the environment and our data is as simple as a few lines of code.
Participants will be provided with an OpenAI Gym~\cite{gym} wrapper for the environment and a simple interface for loading demonstrations from the \minenet-v0 dataset as illustrated in Figure~\ref{fig:example_code}. 
Our data will be released in the form of Numpy \texttt{.npz} files composed of state-action-reward tuples in vector form,
and can be found along with accompanying documentation on the competition website.

\subsection{Tasks and application scenarios}

\subsubsection{Task}
    
The primary task of the competition is solving the \texttt{ObtainDiamond} environment. 
As previously described (see Section~\ref{sec:data}), agents begin at a random position on a randomly generated Minecraft survival map with no items in their inventory. 
The task consists of controlling an embodied agent to obtain a single diamond.
This task can only be accomplished by navigating the complex item hierarchy of Minecraft. 
The learning algorithm will have direct access to a $64$x$64$ pixel point-of-view observation from the perspective of the embodied Minecraft agent, as well as a set of discrete observations of the agent's inventory for every item required for obtaining a diamond (see Figure~\ref{fig:task_hist}). 
The action space of the agent is the Cartesian product of continuous view adjustment (turning and pitching), binary movement commands (left/right, forward/backward), and discrete actions for placing blocks, crafting items, smelting items, and mining/hitting enemies.  
The agent is rewarded for completing the full task.
Due to the difficulty of the task, the agent is also rewarded for reaching a set of milestones of increasing difficulty that form a set of prerequisites for the full task (see Section~\ref{sec:metrics}).

The competition task embodies two crucial challenges in reinforcement learning: sparse rewards and long time horizons. 
The sparsity of the posed task (in both its time structure and long time horizon) necessitates the use of efficient exploration techniques, human priors for policy bootstrapping, or reward shaping via inverse reinforcement learning techniques. 
Although this task is challenging, preliminary results indicate the potential of existing and new methods utilizing human demonstrations to make progress in solving it (see Section~\ref{sec:baselines}).
    
Progress towards solving the \texttt{ObtainDiamond} environment under strict sample complexity constraints lends itself to the development of sample-efficient--and therefore more computationally accessible--sequential decision making algorithms. 
In particular, because we maintain multiple versions of the dataset and environment for development, validation, and evaluation, it is difficult to engineer domain-specific solutions to the competition challenge. 
The best performing techniques must explicitly implement strategies that efficiently leverage human priors across general domains. 
In this sense, the application scenarios of the competition are those which stand to benefit from the development of such algorithms; to that end, we believe that this competition is a step towards democratizing access to deep reinforcement learning based techniques and enabling their application to real-world problems.
    
\paragraph{Previous Year's Task} Stability in metrics across years is crucial for tracking and assessing long-term impact and progress. In the MineRL 2019 competition, no team obtained a diamond; however, many teams made great progress toward solving this task. In fact, the top team was able to obtain the penultimate item to the goal. For this reason, we elected to keep the same task from last year.

\newpage

\subsection{Metrics}
\label{sec:metrics}

\begin{wraptable}{r}{0.5 \textwidth}
    \vspace{-20pt}
    \centering
    \tiny
    \begin{tabular}{ll|ll} 
        {Milestone} & {Reward} & {Milestone} & {Reward} \\
        \midrule
        log & 1                & furnace & 32  \\
        planks & 2             & stone\_pickaxe & 32  \\
        stick & 4              & iron\_ore & 64  \\
        crafting\_table & 4    & iron\_ingot & 128  \\
        wooden\_pickaxe & 8    & iron\_pickaxe &  256 \\
        stone & 16             & diamond & 1024 
     \end{tabular}
     \caption{
        \small Rewards for sub-goals and main goal (diamond) for \texttt{Obtain Diamond}.}
    \label{table:rew}
    \vspace{-10pt}
\end{wraptable} 

Following training, participants will be evaluated on the average score of their model over 500 episodes.
Scores are computed as the sum of the milestone rewards achieved by the agent in a given episode as outlined in Table~\ref{table:rew}. 
A milestone is reached when an agent obtains the first instance of the specified item. 
Ties are broken by the number of episodes required to achieve the last milestone. 
An automatic evaluation script will be included with starter code. 
For official evaluation and validation, a fixed map seed will be selected for each episode. These seeds will not be available to participants during the competition.

\subsection{Baselines, Code, and Material Provided} \label{sec:baselines}

\paragraph{Preliminary Baselines}

\begin{figure}
    \begin{center}

        \includegraphics[width=0.49\textwidth]{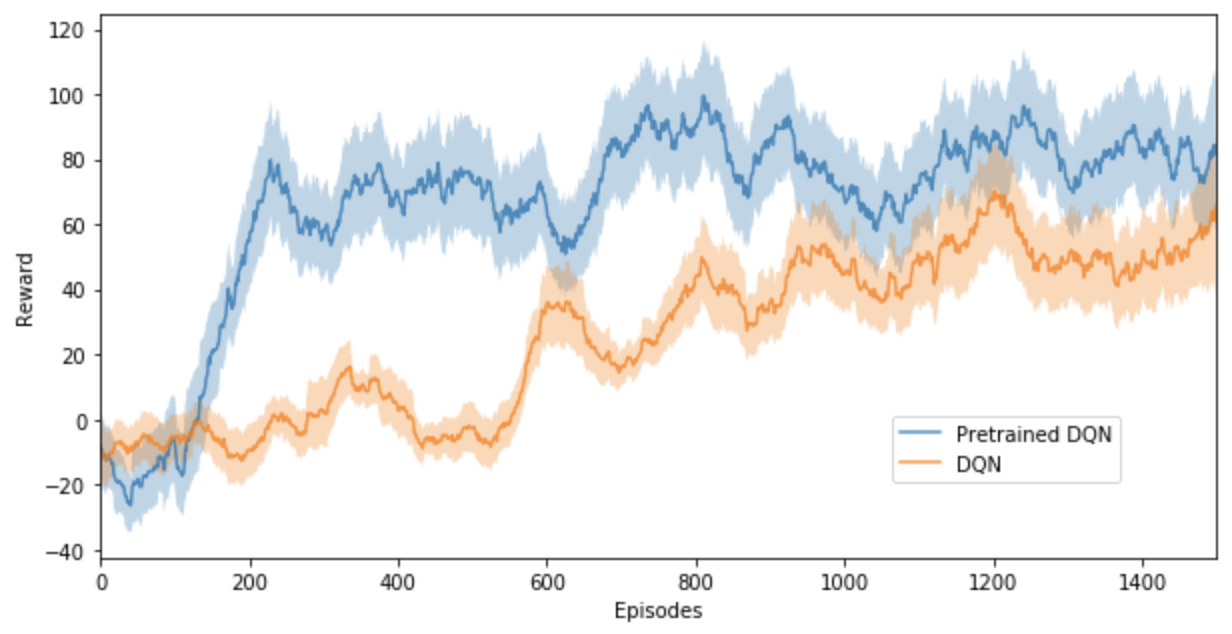} 
        \caption{\small Performance graphs over time with DQN and PreDQN on \texttt{Navigate}(Dense)}
        \label{fig:dqn}

    \end{center}
\end{figure}

We present preliminary results showing the usefulness of the data for improving sample efficiency and overall performance. 
We compare algorithms by the highest average reward obtained over a 100-episode window during training.
We also report the performance of random policies and 50th percentile
human performance.
The results are summarized in Table~\ref{table:perf}.

In the presented comparison, DQN is an implementation of Double Dueling DQN~\cite{hasselt2015deep} and Behavioral Cloning is a supervised learning method trained on expert trajectories.
PreDQN denotes a version of DQN pretrained on the \minenet{}-v0 data: specifically, PreDQN is trained by performing Bellman updates on minibatches drawn from expert trajectories with accompanying reward labels. Before training, we initialize the replay buffer with expert demonstrations.

In all environments, the learned agents perform significantly worse than humans.
 \texttt{Treechop} exhibits the largest difference: on average, humans achieve a score of 64, but reinforcement agents achieve scores of less than 4. 
These results suggest that our environments are quite challenging, especially given that the \texttt{Obtain<Item>} environments build upon the Treechop environment by requiring the completion of several additional sub-goals.
We hypothesize that a large source of difficulty stems from the environment’s inherent
long-horizon credit assignment problems.
For example, it is hard for agents to learn to navigate through water because it
takes many transitions before the agent dies by drowning.

In light of these difficulties, our data is useful in improving performance and sample efficiency:
in all environments, methods that leverage human data perform better.
As seen in Figure~\ref{fig:dqn}, the expert demonstrations were able to achieve higher
reward per episode and attain high performance using fewer samples.
Expert demonstrations are particularly helpful in environments where random exploration 
is unlikely to yield any reward, like Navigate (Sparse). These preliminary results indicate that human demonstrations will be crucial in solving the main competition environment.

\begin{table}
    \small
    \centering
            \begin{tabular}{lccc}
                \toprule
                  & \texttt{Treechop} & \texttt{Navigate} (S) & \texttt{Navigate} (D) \\
                  \midrule
                 DQN \cite{mnih2015human}& 3.73 $\pm$ 0.61 & 0.00 $\pm$ 0.00 & 55.59 $\pm$ 11.38 \\
                 A2C \cite{mnih2016asynchronous}& 2.61 $\pm$ 0.50 & 0.00 $\pm$ 0.00 & -0.97 $\pm$ 3.23 \\
                 Behavioral Cloning & \textbf{43.9} $\pm$ \textbf{31.46} & 4.23 $\pm$ 4.15 & 5.57 $\pm$ 6.00 \\
                PreDQN & {4.16} $\pm$ {0.82} & {6.00} $\pm$ \textbf{4.65} & \textbf{94.96} $\pm$ \textbf{13.42} \\
                \midrule
                Human & 64.00 $\pm$ 0.00 & 100.00 $\pm$ 0.00 & 164.00 $\pm$ 0.00 \\
                Random & 3.81 $\pm$ 0.57 & 1.00 $\pm$ 1.95 & -4.37 $\pm$ 5.10 \\
                \bottomrule
            \end{tabular}
    \caption{
        \small Results in \texttt{Treechop}, \texttt{Navigate} (S)parse, and \texttt{Navigate} (D)ense, over the best 100 contiguous episodes. $\pm$ denotes standard deviation. 
        Note: humans achieve the maximum score for all environments shown.
    }
    \label{table:perf}
    \vspace{0pt}
\end{table}

\paragraph{2019 Baselines} 
For the 2019 MineRL Competition, Preferred Networks\footnote{\url{https://preferred.jp/en/}} provided extensive baselines\footnote{\url{https://github.com/minerllabs/baselines}}, including behavioral cloning, deep Q-learning from demonstrations (DQfD)~\citep{hester2018deep}, Rainbow~\citep{hessel2018rainbow}, generative adversarial inverse RL (GAIL)~\citep{gail_2016}, and proximal policy optimization (PPO)~\citep{ppo}. 
These baselines are implemented using ChainerRL~\cite{fujita2019chainerrl}, and MineRL 2019 participants found them to be incredibly helpful for developing their algorithms.
These baselines\footnote{Preferred Network's writeup of their experiments using their baselines and the MineRL environments can be found~\href{https://github.com/minerllabs/baselines/tree/master/general/chainerrl}{here}.} are available to participants to freely use in this iteration of the competition.

\paragraph{2020 Baselines}
We have again partnered with Preferred Networks to produce high-quality baselines. This year, the baselines are implemented using PyTorch~\cite{pytorch}.
These baselines consist of state-of-the-art RL and imitation learning algorithms, including Rainbow, SQIL~\cite{reddy2019sqil}, prioritized dueling double DQN (PDF DQN)~\cite{schaul2015prioritized,van2016deep,wang2016dueling}, and DQfD.
These baselines fully comply with the rules of this year's competition.

In addition to these baselines, we provide the code from the 2019 baselines and the top teams of the MineRL 2019 competition. 
However, these solutions do not conform to the rules of this year.
We hope competitors will be able to take inspiration from these methods.

\paragraph{Starting Code and Documentation.}
We released an open-source Github repository with starting code including the baselines mentioned above,
an OpenAI Gym interface for the Minecraft simulator, and a data-loader to accompany the data. 
Additionally, we released a public Docker container for ease of use.
We also provide participants with the code for the solutions from last year's top participants.

\subsection{Tutorial and documentation}

We have a competition page that contains instructions, documentation\footnote{\url{http://minerl.io/docs/}}, and updates to the competition. 
For this competition, we plan to include a step-by-step demonstration showing participants how to submit their learning procedures.
Although top participants in MineRL 2019 stated that they found the documentation to be helpful, we plan to extend the documentation in the hopes that even more people can participate this year.

\section{Organizational Aspects}

\subsection{Protocol}

\subsubsection{Submission Protocol}
\label{subsec:sub_protocol}
The evaluation of the submissions will be managed by AIcrowd, an open-source platform for organizing machine learning competitions. 
Throughout the competition, participants will work on their code bases as git repositories\footnote{\url{https://gitlab.aicrowd.com}}. 
Participants must package their intended runtime in their repositories to ensure that the AIcrowd evaluators can automatically build relevant Docker images from their repositories and orchestrate them as needed. 
This approach also ensures that all successfully-evaluated, user-submitted code is both versioned across time and completely reproducible. 

\paragraph{Software Runtime Packaging.}
Packaging and specification of the software runtime is among the most time consuming (and frustrating) task for many participants. 
To simplify this step, we will support numerous approaches to package the software runtime with the help of \texttt{aicrowd-repo2docker}\footnote{\url{https://pypi.org/project/aicrowd-repo2docker/}}. 
The \texttt{aicrowd-repo2docker} is a tool which lets participants specify their runtime using Anaconda environment exports, \texttt{requirements.txt}, or a traditional Dockerfile. 
This significantly decreases the barrier to entry for less technically-inclined participants by transforming an irritating debug cycle to a deterministic one-liner that performs the work behind the scenes. 

\paragraph{Submission Mechanism.}
Participants will collaborate on their git repository throughout the competition.
Whenever they are ready to make a submission, they will create and push a \textit{git tag} to trigger the evaluation pipeline. 

\paragraph{Orchestration of the Submissions.}
The ability to reliably orchestrate user submissions over large periods of time is a key determining feature of the success of the proposed competition. 
We will use the evaluators of AIcrowd, which use custom Kubernetes clusters to orchestrate the submissions against pre-agreed resource usage constraints. 
The same setup has previously been successfully used in numerous other machine learning competitions, such as NeurIPS 2017: Learning to Run Challenge, NeurIPS 2018: AI for Prosthetics Challenge, NeurIPS 2018: Adversarial Vision Challenge, and the 2018 MarLO challenge.

\subsubsection{General Competition Structure}

\paragraph{Round 1: General Entry.} 
In this round, participants will register on the competition website, and receive the following materials:
\begin{itemize}
	\item Starter code for running the MineRL environments for the competition task. 
	\item Basic baseline implementations provided by Preferred Networks, the competition organizers, and the top teams from the MineRL 2019 competition (see Section~\ref{sec:baselines}).
	\item Two different renders of the human demonstration dataset (one for methods development, the other for validation) with modified textures, lighting conditions, and minor game state changes. 
	\item The Docker Images and quick-start template that the competition organizers will use to validate the training performance of the competitor's models.
\end{itemize}

Competitors may submit solutions to two tracks. The main track will provide access to both the simulator and paired demonstrations during training, while the alternate, demonstrations only track, will only provide agents with the MineRL-v0 dataset during training. Both tracks will be evaluated by measuring average performance over 100 episodes on the \texttt{ObtainDiamond} task. Competitors may submit to both tracks.

When satisfied with their models, participants will follow the submission protocols (described in Section~\ref{subsec:sub_protocol}) to submit their code for evaluation, specifying either the main track or alternate track. 
The automated evaluation setup will evaluate the submissions against the validation environment, to compute and report the metrics (described in Section~\ref{sec:metrics}) to the respective public leaderboard on the AIcrowd website. 
Because the full training phase is quite resource intensive, it is not be possible to run the training for all the submissions in this round; however, the evaluator will ensure that the submitted code includes the relevant subroutines for the training of the models by running a short integration test on the training code before doing the actual evaluation on the validation environment.

Once Round 1 is complete, the organizers will examine the code repositories of the top submissions from each track to ensure compliance with the competition rules. For the main track, the top 15 verified teams will be invited to the second round. For the alternate demonstrations-only track, 5 teams will move on to Round 2. To verify the top submissions comply with the competition rules, they will be automatically trained on the validation dataset and environment by the competition orchestration platform. The code repositories associated with the corresponding submissions will be forked, and scrubbed of large files to ensure that participants are not using any pretrained models in the subsequent round.   
The resulting trained models will then be evaluated over several hundred episodes.
Their performance will be compared with the submission's final model performance during Round 1 to ensure that no warm-starting or adversarial modifications of the evaluation harness was made. In the case of the demonstrations-only track, we additionally verify that no environment interactions were used in the development of the model.
The teams whose submissions have conflicting end-of-round and organizer-ran performance distribution will be contacted for appeal. 
Unless a successful appeal is made, the organizers will remove those submissions from the competition and then evaluate additional submissions until each track is at capacity: 15 teams for the main track, and 5 teams for the alternate track. Teams may qualify for the second round in both tracks; therefore, fewer than 20 teams may qualify for Round 2 among the two tracks.

\paragraph{Round 2: Finals.} 
In this round, the top performing teams will continue to develop their algorithms.
Their work will be evaluated against a confidential, held-out test environment and test dataset, to which they will not have access. This environment includes perturbations to both the action-space as well as the observation space.

Specifically, participants in each track will be able to make a submission to that track (as described in Section~\ref{subsec:sub_protocol}) twice during Round 2. The automated evaluator will execute their algorithms on the test dataset and simulator, and report their score and metrics back to the participants. This is done to prevent competitors from over-fitting to the training and validation datasets/simulators.

Again all submitted code repositories will be scrubbed to remove any files larger than 15MB to ensure participants are not including any model weights pre-trained on the previously released training dataset. While the container running the submitted code will not have external network access, relevant exceptions are added to ensure participants can download and use popular frameworks like PyTorch\footnote{\url{https://pytorch.org}} and Tensorflow\footnote{\url{http://tensorflow.org}}. Participants can request to add network exceptions for any other publicly available resource, which will be validated by AIcrowd on a case by case basis.

Further, participants will submit a written report of their technical approach to the problem; this report will be used to bolster the impact of this competition on sample-efficient reinforcement learning research. They will also be encouraged to submit their papers to relevant workshops at NeurIPS in order to increase interest in their work.

At the end of the second period, the competition organizers will execute a final run of the participants’ algorithms and the winners will be selected for each of the competition tracks. 

\paragraph{User Submitted Code.} 
If a team requires an exception to the open source policy, then the team has a time window of 3 weeks after the competition ends to request an appeal by contacting the organizers. We will communicate with the team and potentially grant an exception. For example, a submission may be open sourced at a later date if the team is preparing a research publication based on new techniques used within their submission.
By default, all of the associated code repositories will be made public and available\footnote{\url{https://gitlab.aicrowd.com}} after the 3 week window at the end of the competition.

\paragraph{NeurIPS Workshop.} 

After winners have been selected, there will be a NeurIPS workshop to exhibit the technical approaches developed during the competition.
We plan to invite teams from Round 2 to attend and present their results at the workshop.
Due to COVID-19, this workshop will be completely online.

\subsection{Rules}
The aim of the competition is to develop sample-efficient training algorithms. Therefore, we discourage the use of
environment-specific, hand-engineered features because they do not demonstrate fundamental algorithmic improvements.
The following rules attempt to capture the spirit of the competition and any submissions found to be violating 
the rules may be deemed ineligible for participation by the organizers.
\begin{itemize}
	\item \textbf{Entries to the MineRL competition must be ``open''.} Teams will be expected to reveal most details of their method including source-code (special exceptions may be made for pending publications).
	\item \textbf{For a team to be eligible to move to Round 2, each member must satisfy the following conditions:}
	\begin{itemize}
	    \item be at least 18 and at least the age of majority in place of residence; 
	    \item not reside in any region or country subject to U.S. Export Regulations; and 
	    \item not be an organizer of this competition nor a family member of a
	    competition organizer. 
	    \end{itemize}
    \item \textbf{To receive any awards from our sponsors, competition winners must attend the NeurIPS workshop.}
    \item \textbf{The submission must train a machine learning model without relying on human domain knowledge.}
    \begin{itemize}
        \item \textbf{The reward function may not be changed (shaped) based on manually engineered, hard-coded functions of the state.} For example, additional rewards for approaching tree-like objects are not permitted, but rewards for encountering novel states (“curiosity rewards”) are permitted.
        \item \textbf{Actions/meta-actions/sub-actions/sub-policies may not be manually specified in any way.} For example, though a learned hierarchical controller is permitted, meta-controllers may not choose between two policies based on a manually specified condition, such as whether the agent has a certain item in its inventory. This restriction includes the composition of actions (e.g., adding an additional action which is equivalent to performing “walk forward for 2 seconds” or “break a log and then place a crafting table”).
        \item \textbf{State processing/pre-processing cannot be hard-coded with the exception of frame-stacking.} For example, the agent can act every even-numbered timestep based on the last two observations, but a manually specified edge detector may not be applied to the observation. As another example, the agent’s observations may be normalized to be “zero-mean, variance one” based on an observation history or the dataset.
        \item \textbf{To ensure that the semantic meaning attached to action and observation labels are not exploited, the labels assigned to actions and observations have been obfuscated (in both the dataset and the environment).} Actions and observations (with the exception of POV observations) have been embedded into a different space. Furthermore, during Round 2 submissions, the actions will be re-embedded. Any attempt to bypass these obfuscations will constitute a violation of the rules.
        \item \textbf{Models may only be trained against the competition environments (MineRL environments ending with “VectorOb(f)").}  All of the MineRL environments have specific competition versions which incorporate action and observation space obfuscation. They all share a similar observation and action space embedding which is changed in Round 2 as with the texture pack of the environment. 
    \end{itemize}
    \item \textbf{There are two tracks, each with a different sample budget:}
    \begin{itemize}
        \item \textbf{The primary track is ``Demonstrations and Environment.''} Eight million (8,000,000) interactions with the environment may be used in addition to the provided dataset. If stacking observations / repeating actions, then each skipped frame still counts against this budget.
        \item \textbf{The secondary track is ``Demonstrations Only.''} No environment interactions may be used in addition to the provided dataset. Competitors interested in learning solely from demonstrations can compete in this track without being disadvantaged compared to those who also use reinforcement learning. 
        \item \textbf{A team can submit separate entries to both tracks; performance in the tracks will be evaluated separately} (i.e., submissions between the two tracks are not linked in any way).
    \end{itemize}
    \item \textbf{Participants may only use the provided dataset; no additional datasets may be included in the source file submissions nor may be downloaded during training evaluation, but pre-trained models which are publicly available by June 5th are permitted.}
    \begin{itemize}
        \item During the evaluation of submitted code, the individual containers will \textit{not} have access to any external network in order to avoid any information leak. Relevant exceptions are added to ensure participants can download and use the pre-trained models included in popular frameworks like PyTorch and TensorFlow. Participants can request to add network exceptions for any other publicly available pre-trained models, which will be validated by AICrowd on a case-by-case basis.
        \item All submitted code repositories will be scrubbed to remove files larger than 30MB to ensure participants are not checking in any model weights pretrained on the released training dataset.
        \item Pretrained models are not allowed to have been trained on MineRL or any related or unrelated Minecraft data. The intent of this rule is to allow participants to use models which are, for example, trained on ImageNet or similar datasets. Don't abuse this.
    \end{itemize}
    \item \textbf{The procedure for Round 1 is as follows:}
    \begin{itemize}
        \item During Round 1, teams submit their trained models for evaluation at most twice a week times and receive the performance of their models. 
        \item At the end of Round 1, teams must submit source code to train their models. This code must terminate within four days on the specified platform. 
        \item For teams with the highest evaluation scores, this code will be inspected for rule compliance and used to re-train the models with the validation dataset and environment. 
        \item For those submissions whose end-of-round and organizer-ran performance distributions disagree, the offending teams will be contacted for appeal. Unless a successful appeal is made, the organizers will remove those submissions from the competition and then evaluate additional submissions until each track is at capacity.
        \item The top 15 teams in the main (RL+Demonstration) track and the top 5 teams in the secondary (Demonstration Only) track will progress to Round 2.
    \end{itemize}   
    \item \textbf{The procedure for Round 2 is as follows:}
    \begin{itemize}
        \item During Round 2, teams will submit their source code at most once every two weeks.
        \item After each submission, the model will be trained for four days on a re-rendered, private dataset and domain, and the teams will receive the final performance of their model. The dataset and domain will contain matching perturbations to the action space and the observation space.
        \item At the end of the round, final standings are based on the best-performing submission of each team during Round 2.
    \end{itemize}
    \item \textbf{Official rule clarifications will be made in the FAQ on the AIcrowd website.}
    \begin{itemize}
    \item The FAQ is available here\footnote{\url{https://www.aicrowd.com/challenges/neurips-2020-minerl-competition\#faq}}.

    \item Answers within the FAQ are \textit{official answers} to questions. Any informal answers to questions (e.g., via email) are superseded by answers added to the FAQ.
\end{itemize}

\end{itemize}

See the rules page\footnote{\url{https://www.aicrowd.com/challenges/neurips-2020-minerl-competition/challenge_rules}} (an AIcrowd account is needed to view this page) for any updates.

\paragraph{Cheating.}
The competition is designed to prevent rule breaking and to discourage submissions that circumvent the competition goals.
Submissions will be tested on variants of the environment/data with different textures and lighting,
discouraging the any priors that are not trained from scratch. 
Inherent stochasticity in the environment, such as different world and spawn locations, as well as the desemantization and isomorphic embedding of state and action-space components directly discourage the use of hard-coded policies. 
Furthermore, we will use automatic evaluation scripts to verify the participants' submitted scores in the first round
and perform a manual code review of the finalists of each round in the competition. We highlight that the evaluation dataset/environment pair on which participants will be evaluated is \emph{completely inaccessible} to competitors, and measures are taken to prevent information leak.

\subsection{Schedule and Readiness}

\subsubsection{Schedule}  

Given the difficulty of the problem posed, ample time shall be given to allow participants to fully realize their solutions.
Our proposed timeline gives competitors over 80 days to prepare, evaluate, and receive feedback on their solutions before the end of the first round.
\begin{itemize}
	\item [April 13] \textbf{Competition Accepted}. 
	\item [May] \textbf{Pre-Release:} Submission framework finalized. 
	\item [June] \textbf{First Round Begins:} Participants invited to download starting materials and baselines and to begin developing their submission.
	\item [September] \textbf{End of First Round:} Submissions close. Models evaluated by organizers and partners.
	\item [September] \textbf{First Round Results Posted:} Official results posted notifying finalists.
	\item [September] \textbf{Final Round Begins:} Finalists invited to submit their models against the held out validation texture pack.
	\item [November] \textbf{End of Final Round:} Submissions close. Organizers train finalists latest submission for evaluation.
	\item [November] \textbf{Final Results Posted:} Official results of model training and evaluation posted.
	\item [December 6] \textbf{NeurIPS 2020:} Winning teams invited to the conference to present their results. Awards announced at conference.
\end{itemize}

\subsubsection{Readiness.} At the time of writing this proposal the following key milestones are complete: 
\begin{itemize}
    \item The dataset is fully collected, cleaned, and automatically annotated;
    \item The competition environments have been finalized and implemented;
    \item The advisory committee is fully established; 
    \item The partnership with AIcrowd has been confirmed, and we are in discussion with last year's sponsors;
    \item A specific plan for attracting underrepresented groups is finalized; 
    \item The competition infrastructure has been developed, including the submission harness. 
\end{itemize}
If accepted to the NeurIPS competition track, there are no major roadblocks preventing the execution of the competition.

\subsection{Competition promotion}

\paragraph{Partnership with Affinity Groups}
We hope to partner with affinity groups to promote the participation of groups who are traditionally underrepresented at NeurIPS.
We plan to reach out to organizers of Women in Machine Learning (WiML)\footnote{\url{https://wimlworkshop.org/}}, LatinX in AI (LXAI)\footnote{\url{https://www.latinxinai.org/}}, Black in AI (BAI)\footnote{\url{https://blackinai.github.io/}}, and Queer in AI\footnote{\url{https://sites.google.com/view/queer-in-ai/}}. 
We will also reach out to organizations, such as Deep Learning Indaba\footnote{\url{http://www.deeplearningindaba.com/}} and Data Science Africa\footnote{\url{http://www.datascienceafrica.org/}}, to determine how to increase the participation of more diverse teams.
Specifically, we hope to form a selection committee for the Inclusion@NeurIPS scholarships consisting of some of our organizers and members from those groups.
We also plan to encourage competition participants to submit write-ups of their solutions to relevant affinity group workshops at NeurIPS.

\paragraph{Promotion through General Mailing Lists}
To promote participation in the competition, we plan to distribute the call to general technical mailing lists, such as Robotics Worldwide and Machine Learning News; company mailing lists, such as DeepMind's internal mailing list; and institutional mailing lists. 
We plan to promote participation of underrepresented groups in the competition by distributing the call to affinity group mailing lists, including, but not limited to Women in Machine Learning, LatinX in AI, Black in AI, and Queer in AI.
Furthermore, we will reach out to individuals at historically black or all-female universities and colleges to encourage the participation of these students and/or researchers in the competition.
By doing so, we will promote the competition to individuals who are not on any of the aforementioned mailing lists, but are still members of underrepresented groups. 

\paragraph{Media Coverage}
To increase general interest and excitement surrounding the competition, we will reach out to the media coordinator at Carnegie Mellon University.
By doing so, our competition will be promoted by popular online magazines and websites, such as Wired. 
We will also post about the competition on relevant popular subreddits, such as \url{r/machinelearning} and /r/datascience, and promote it through social media. 
We will utilize our industry and academic partners to post on their various social media platforms, such as the OpenAI Blog, the Carnegie Mellon University Twitter, and the Microsoft Facebook page.

The previous iteration of the MineRL competition was featured by several notable news outlets including Nature News~\cite{hsu_2019}, BBC~\cite{shead_2019}, The Verge~\cite{vincent_2019}, and Synced~\cite{synced_2019}. This widespread publication and coverage of the competition led to a drastic influx of new users and spectators from outside of the NeurIPS community. We intend on further leveraging these media connections to increase the reach of our call for competitors.

\section{Resources}

\subsection{Organizing team}

\subsubsection{Organizers}

\paragraph{William H. Guss.} William Guss is a research scientist at OpenAI and Ph.D. student in the Machine Learning Department at CMU. William co-created the \minenet{} dataset and lead the MineRL  competition at NeurIPS 2019. He is advised by Dr. Ruslan Salakhutdinov
and his research spans sample-efficient reinforcement learning and deep learning theory.  William completed his bachelors in Pure Mathematics at UC Berkeley where he was awarded the Regents' and Chancellor's Scholarship, the highest honor awarded to incoming undergraduates. During his time at Berkeley, William received the Amazon Alexa Prize Grant for the development of conversational AI and co-founded Machine Learning at Berkeley. William is from Salt Lake City, Utah and grew up in an economically impacted, low-income neighborhood without basic access to computational resources. As a result, William is committed to working towards developing research and initiatives which promote socioeconomically-equal access to AI/ML systems and their development.

\paragraph{Mario Ynocente Castro.} Mario is an Engineer at Preferred Networks. In 2017, he received a Masters in Applied Mathematics at École polytechnique and a Masters in Machine Learning at École Normal Supérieure de Paris-Saclay. His current work focuses on applications of Reinforcement Learning and Imitation Learning.

\paragraph{Sam Devlin.} Sam Devlin is a Senior Researcher in the Game Intelligence and Reinforcement Learning research groups at Microsoft Research, Cambridge (UK). He received his PhD on multi-agent reinforcement learning in 2013 from the University of York. Sam has previously co-organised the Text-Based Adventure AI Competition in 2016 \& 2017 and the Multi-Agent Reinforcement Learning in Minecraft (MARLO) Competition in 2018.

\paragraph{Brandon Houghton.} Brandon Houghton is a Machine Learning Engineer at OpenAI and co-creator of the \minenet{} dataset. Graduating from the School of Computer Science at Carnegie Mellon University, Brandon's work focuses on developing techniques to enable agents to interact with the real world through virtual sandbox worlds such as Minecraft. He has worked on many machine learning projects, such as discovering model invariants in physical systems as well as learning lane boundaries for autonomous driving. 

\paragraph{Noboru Sean Kuno.} Noboru Sean Kuno is a Senior Research Program Manager at Microsoft Research in Redmond, USA. He is a member of Artificial Intelligence Engaged team of Microsoft Research Outreach. He leads the design, launch and development of research programs for AI projects such as Project Malmo, working in partnership with research communities and universities worldwide.

\paragraph{Crissman Loomis.} Crissman works for Preferred Networks, a Japanese AI startup that applies the latest deep machine learning algorithms to industrial applications, like self-driving cars, factory automation, or medicine development. At Preferred Networks, he has supported the development and adoption of open source frameworks, including the deep learning framework Chainer and more recently the hyperparameter optimization library Optuna.

\paragraph{Stephanie Milani.} Stephanie Milani is a Ph.D. student in the Machine Learning Department at Carnegie Mellon University.
She is advised by Dr. Fei Fang and her research interests include sequential decision-making problems, with an emphasis on reinforcement learning.
In 2019, she completed her B.S. in Computer Science and her B.A. in Psychology at the University of Maryland, Baltimore County, and she co-organized the 2019 MineRL competition..
Since 2016, she has worked to increase the participation of underrepresented groups in CS and AI at the local and state level. 
For these efforts, she has been nationally recognized 
through a Newman Civic Fellowship. 

\paragraph{Sharada Mohanty.} Sharada Mohanty is the CEO and Co-founder of AIcrowd, an open-source platform encouraging reproducible artificial intelligence research. 
He was the co-organizer of many large-scale machine learning competitions, such as NeurIPS 2017: Learning to Run Challenge, NeurIPS 2018: AI for Prosthetics Challenge, NeurIPS 2018: Adversarial Vision Challenge, NeurIPS 2019 : MineRL Competition, NeurIPS 2019: Disentanglement Challenge, NeurIPS 2019: REAL Robots Challenge. 
During his Ph.D. at EPFL, he worked on numerous problems at the intersection of AI and health, with a strong interest in reinforcement learning.  
In his current role, he focuses on building better engineering tools for AI researchers and making research in AI accessible to a larger community of engineers. 

\paragraph{Keisuke Nakata.} Keisuke Nakata is a machine learning engineer at Preferred Networks, Inc. He mainly works on machine learning applications in real-world industry settings. Particularly, his interests lie in creating reinforcement learning algorithms and frameworks.

\paragraph{Ruslan Salakhutdinov.} Ruslan Salakhutdinov received his Ph.D. in machine learning (computer science) from the University of Toronto in 2009. After spending two post-doctoral years at the Massachusetts Institute of Technology Artificial Intelligence Lab, he joined the University of Toronto as an Assistant Professor in the Department of Computer Science and Department of Statistics. In February of 2016, he joined the Machine Learning Department at Carnegie Mellon University as an Associate Professor. Ruslan's primary interests lie in deep learning, machine learning, and large-scale optimization. His main research goal is to understand the computational and statistical principles required for discovering structure in large amounts of data. He is an action editor of the Journal of Machine Learning Research and served on the senior programme committee of several learning conferences including NeurIPS and ICML. He is an Alfred P. Sloan Research Fellow, Microsoft Research Faculty Fellow, Canada Research Chair in Statistical Machine Learning, a recipient of the Early Researcher Award, Connaught New Researcher Award, Google Faculty Award, Nvidia's Pioneers of AI award, and is a Senior Fellow of the Canadian Institute for Advanced Research.

\paragraph{John Schulman.} John Schulman is a researcher and founding member of OpenAI, where he leads the reinforcement learning team. He received a PhD from UC Berkeley in 2016, advised by Pieter Abbeel. He was named one of MIT Tech Review’s 35 Innovators Under 35 in 2016.

\paragraph{Shinya Shiroshita.} 
Shinya Shiroshita works for Preferred Networks as an engineer. He graduated from the University of Tokyo, where he majored in computer science. His hobbies are competitive programming and playing board games. In Minecraft, he likes exploring interesting structures and biomes.

\paragraph{Nicholay Topin.} Nicholay Topin is a Machine Learning Ph.D. student advised by Dr. Manuela Veloso at Carnegie Mellon University. His current research focus is explainable deep reinforcement learning systems. Previously, he has worked on knowledge transfer for reinforcement learning and learning acceleration for deep learning architectures. 

\paragraph{Avinash Ummadisingu.} Avinash Ummadisingu works at Preferred Networks on Deep Reinforcement Learning for Robotic Manipulation and the open-source library PFRL (formerly ChainerRL). His areas of interests include building sample efficient reinforcement learning systems and multi-task learning. Prior to that, he was a student at USI, Lugano under the supervision of Prof. Jürgen Schmidhuber and Dr. Paulo E. Rauber of the Swiss AI Lab IDSIA.

\paragraph{Oriol Vinyals.} Oriol Vinyals is a Principal Scientist at Google DeepMind, and a team lead of the Deep Learning group. His work focuses on Deep Learning and Artificial Intelligence. Prior to joining DeepMind, Oriol was part of the Google Brain team. He holds a Ph.D. in EECS from the University of California, Berkeley and is a recipient of the 2016 MIT TR35 innovator award. His research has been featured multiple times at the New York Times, Financial Times, WIRED, BBC, etc., and his articles have been cited over 65000 times. His academic involvement includes program chair for the International Conference on Learning Representations (ICLR) of 2017, and 2018. He has also been an area chair for many editions of the NIPS and ICML conferences. Some of his contributions such as seq2seq, knowledge distillation, or TensorFlow are used in Google Translate, Text-To-Speech, and Speech recognition, serving billions of queries every day, and he was the lead researcher of the AlphaStar project, creating an agent that defeated a top professional at the game of StarCraft, achieving Grandmaster level, also featured as the cover of Nature. At DeepMind he continues working on his areas of interest, which include artificial intelligence, with particular emphasis on machine learning, deep learning and reinforcement learning.

\subsubsection{Advisors}
\paragraph{Anca Dragan.} Anca Dragan is an Assistant Professor in the EECS Department at UC Berkeley. Her goal is to enable robots to work with, around, and in support of people. She runs the InterACT Lab, where the focus is on algorithms for human-robot interaction -- algorithms that move beyond the robot's function in isolation, and generate robot behavior that also accounts for interaction and coordination with end-users. The lab works across different applications, from assistive robots, to manufacturing, to autonomous cars, and draw from optimal control, planning, estimation, learning, and cognitive science. She also helped found and serve on the steering committee for the Berkeley AI Research (BAIR) Lab, and am a co-PI of the Center for Human-Compatible AI. She was also honored by the Sloan Fellowship, MIT TR35, the Okawa award, and an NSF CAREER award.

\paragraph{Fei Fang.} Fei Fang is an Assistant Professor at the Institute for Software Research in the School of Computer Science at Carnegie Mellon University. Before joining CMU, she was a Postdoctoral Fellow at the Center for Research on Computation and Society (CRCS) at Harvard University. She received her Ph.D. from the Department of Computer Science at the University of Southern California in June 2016.
Her research lies in the field of artificial intelligence and multi-agent systems, focusing on integrating machine learning with game theory. Her work has been motivated by and applied to security, sustainability, and mobility domains, contributing to the theme of AI for Social Good.

\paragraph{Chelsea Finn.} Chelsea Finn is an Assistant Professor in Computer Science and Electrical Engineering at Stanford University. Finn's research interests lie in the capability of robots and other agents to develop broadly intelligent behavior through learning and interaction. To this end, her work has included deep learning algorithms for concurrently learning visual perception and control in robotic manipulation skills, inverse reinforcement methods for scalable acquisition of nonlinear reward functions, and meta-learning algorithms that can enable fast, few-shot adaptation in both visual perception and deep reinforcement learning. Finn received her Bachelor's degree in Electrical Engineering and Computer Science at MIT and her PhD in Computer Science at UC Berkeley. Her research has been recognized through the ACM doctoral dissertation award, an NSF graduate fellowship, a Facebook fellowship, the C.V. Ramamoorthy Distinguished Research Award, and the MIT Technology Review 35 under 35 Award, and her work has been covered by various media outlets, including the New York Times, Wired, and Bloomberg. Throughout her career, she has sought to increase the representation of underrepresented minorities within CS and AI by developing an AI outreach camp at Berkeley for underprivileged high school students, a mentoring program for underrepresented undergraduates across four universities, and leading efforts within the WiML and Berkeley WiCSE communities of women researchers.

\paragraph{David Ha.} David Ha 
is a Research Scientist at Google Brain. His research interests include Recurrent Neural Networks, Creative AI, and Evolutionary Computing. Prior to joining Google, He worked at Goldman Sachs as a Managing Director, where he co-ran the fixed-income trading business in Japan. He obtained undergraduate and graduate degrees in Engineering Science and Applied Math from the University of Toronto. 
 
\paragraph{Sergey Levine.} Sergey Levine received a BS and MS in Computer Science from Stanford University in 2009, and a Ph.D. in Computer Science from Stanford University in 2014. He joined the faculty of the Department of Electrical Engineering and Computer Sciences at UC Berkeley in fall 2016. His work focuses on machine learning for decision making and control, with an emphasis on deep learning and reinforcement learning algorithms. Applications of his work include autonomous robots and vehicles, as well as computer vision and graphics. He has previously served as the general chair for the Conference on Robot Learning, program co-chair for the International Conference on Learning Representations, and organizer for numerous workshops at ICML, NeurIPS, and RSS. He has also served as co-organizer on the \emph{Learning to Run} and \emph{AI for Prosthetics} NeurIPS competitions.

\paragraph{Zachary Chase Lipton.} 
Zachary Chase Lipton is an assistant professor of Operations Research and Machine Learning at Carnegie Mellon University. His research spans core machine learning methods and their social impact and addresses diverse application areas, including clinical medicine and natural language processing. Current research focuses include robustness under distribution shift, breast cancer screening, the effective and equitable allocation of organs, and the intersection of causal thinking and the messy high-dimensional data that characterizes modern deep learning applications. He is the founder of the Approximately Correct blog (approximatelycorrect.com) and a founder and co-author of Dive Into Deep Learning, an interactive open-source book drafted entirely through Jupyter notebooks.

\paragraph{Manuela Veloso.} Manuela Veloso is a Herbert A. Simon University Professor at Carnegie Mellon University and the head of AI research at JPMorgan Chase.
She received her Ph.D. in computer science from Carnegie Mellon University in 1992.
Since then, she has been a faculty member at the Carnegie Mellon School of Computer Science.
Her research focuses on artificial intelligence and robotics, across a range of planning, execution, and learning algorithms.
She cofounded the RoboCup Federation and served as president of AAAI from 2011 to 2016.
She is a AAAI, IEEE, AAAS, and ACM fellow.

\subsubsection{Partners and Sponsors}

We are currently in conversation with potential partners for this year's competition.
Last year, we partnered with and/or received support from Microsoft Research, Preferred Networks, NVIDIA, and Artificial Intelligence Journal (AIJ).

\subsection{Resources provided by organizers, including prizes}
{
    \paragraph{Mentorship.}
    We will facilitate a community forum through our publicly available Discord server to enable participants to ask questions, provide feedback, and engage meaningfully with our organizers and advisory board. We hope to foster an active community to collaborate on these hard problems.

    \paragraph{Computing Resources.}
    In concert with our efforts to provide open, democratized access to AI, we are in conversation with potential sponsors to provide compute grants for teams that self identify as lacking access to the necessary compute power to participate in the competition, as we did in the last iteration of the competition.
    We will also provide groups with the evaluation resources for their experiments in Round 2, as we did in the last iteration of the competition.

    \paragraph{Travel Grants and Scholarships.} 
    The competition organizers are committed to increasing the participation of groups traditionally underrepresented in reinforcement learning and, more generally, in machine learning (including, but not limited to: women, LGBTQ individuals, underrepresented racial and ethnic groups, and individuals with disabilities). 
    To that end, we will offer Inclusion@NeurIPS scholarships/travel grants for Round 1 participants who are traditionally underrepresented at NeurIPS to attend the conference. 
    These individuals will be able to apply online for these grants; their applications will be evaluated by the competition organizers and partner affinity groups.
    We also plan to provide travel grants to enable all of the top participants from Round 2 to attend our NeurIPS workshop.
    We are in conversation with potential sponsors about providing funding for these travel grants.

    \paragraph{Prizes.}
    We are currently in discussion about prizes with potential sponsors and/or partners. In the previous competition, we offered 5 NVIDIA GPUs and 10 NVIDIA Jetsons to the top teams. In addition, we provided two prizes for notable research contributions.
}
\subsection{Support and facilities requested}
Due to the quality of sponsorships and industry partnerships secured last year, we only request facility resources and ticket reservations. 
We aim to present at the NeurIPS 2020 Competition Workshop.
We will invite guest speakers, organizers, Round 2 participants, and some Round 1 participants.
To allow these people to attend NeurIPS, we request 30 reservations for NeurIPS. 
We plan to provide funding for teams to travel to the competition.
The organizers will be present at their own expense.

\newpage 

\bibliographystyle{plainnat}
\bibliography{main}

\begin{thebibliography}{44}
\providecommand{\natexlab}[1]{#1}
\providecommand{\url}[1]{\texttt{#1}}
\expandafter\ifx\csname urlstyle\endcsname\relax
  \providecommand{\doi}[1]{doi: #1}\else
  \providecommand{\doi}{doi: \begingroup \urlstyle{rm}\Url}\fi

\bibitem[Amodei and Hernandez(2018)]{amodei_hednandez_2018}
Dario Amodei and Danny Hernandez.
\newblock https://blog.openai.com/ai-and-compute/, May 2018.
\newblock URL \url{https://blog.openai.com/ai-and-compute/}.

\bibitem[Bellemare et~al.(2013)Bellemare, Naddaf, Veness, and
  Bowling]{bellemare2013arcade}
Marc~G Bellemare, Yavar Naddaf, Joel Veness, and Michael Bowling.
\newblock The arcade learning environment: An evaluation platform for general
  agents.
\newblock \emph{Journal of Artificial Intelligence Research}, 47:\penalty0
  253--279, 2013.

\bibitem[Berner et~al.(2019)Berner, Brockman, Chan, Cheung, D{{e}}biak,
  Dennison, Farhi, Fischer, Hashme, Hesse, et~al.]{berner2019dota}
Christopher Berner, Greg Brockman, Brooke Chan, Vicki Cheung, Przemyslaw
  D{{e}}biak, Christy Dennison, David Farhi, Quirin Fischer, Shariq Hashme,
  Chris Hesse, et~al.
\newblock Dota 2 with large scale deep reinforcement learning.
\newblock \emph{arXiv preprint arXiv:1912.06680}, 2019.

\bibitem[Bojarski et~al.(2016)Bojarski, Del~Testa, Dworakowski, Firner, Flepp,
  Goyal, Jackel, Monfort, Muller, Zhang, et~al.]{bojarski2016end}
Mariusz Bojarski, Davide Del~Testa, Daniel Dworakowski, Bernhard Firner, Beat
  Flepp, Prasoon Goyal, Lawrence~D Jackel, Mathew Monfort, Urs Muller, Jiakai
  Zhang, et~al.
\newblock End to end learning for self-driving cars.
\newblock \emph{arXiv preprint arXiv:1604.07316}, 2016.

\bibitem[Brockman et~al.(2016)Brockman, Cheung, Pettersson, Schneider,
  Schulman, Tang, and Zaremba]{gym}
Greg Brockman, Vicki Cheung, Ludwig Pettersson, Jonas Schneider, John Schulman,
  Jie Tang, and Wojciech Zaremba.
\newblock Open{AI} gym.
\newblock \emph{arXiv preprint arXiv:1606.01540}, 2016.

\bibitem[Cruz~Jr et~al.(2017)Cruz~Jr, Du, and Taylor]{cruz2017pre}
Gabriel~V Cruz~Jr, Yunshu Du, and Matthew~E Taylor.
\newblock Pre-training neural networks with human demonstrations for deep
  reinforcement learning.
\newblock \emph{arXiv preprint arXiv:1709.04083}, 2017.

\bibitem[DeepMind(2018)]{deepmind}
DeepMind.
\newblock Alphastar: Mastering the real-time strategy game starcraft ii, 2018.
\newblock URL
  \url{https://deepmind.com/blog/alphastar-mastering-real-time-strategy-game-starcraft-ii/}.

\bibitem[Finn et~al.(2016)Finn, Levine, and Abbeel]{finn2016guided}
Chelsea Finn, Sergey Levine, and Pieter Abbeel.
\newblock Guided cost learning: Deep inverse optimal control via policy
  optimization.
\newblock In \emph{The 33rd International Conference on Machine Learning},
  pages 49--58, 2016.

\bibitem[Finn et~al.(2017)Finn, Yu, Zhang, Abbeel, and Levine]{finn2017one}
Chelsea Finn, Tianhe Yu, Tianhao Zhang, Pieter Abbeel, and Sergey Levine.
\newblock One-shot visual imitation learning via meta-learning.
\newblock \emph{arXiv preprint arXiv:1709.04905}, 2017.

\bibitem[Fujita et~al.(2019)Fujita, Kataoka, Nagarajan, and
  Ishikawa]{fujita2019chainerrl}
Yasuhiro Fujita, Toshiki Kataoka, Prabhat Nagarajan, and Takahiro Ishikawa.
\newblock {ChainerRL}: A deep reinforcement learning library.
\newblock In \emph{The 23rd Conference on Neural Information Processing
  Systems, Deep Reinforcement Learning Workshop}, 2019.

\bibitem[Gao et~al.(2018)Gao, Lin, Yu, Levine, Darrell,
  et~al.]{gao2018reinforcement}
Yang Gao, Ji~Lin, Fisher Yu, Sergey Levine, Trevor Darrell, et~al.
\newblock Reinforcement learning from imperfect demonstrations.
\newblock \emph{arXiv preprint arXiv:1802.05313}, 2018.

\bibitem[Guss et~al.(2019)Guss, Codel*, Hofmann*, Houghton*, Kuno*, Milani*,
  Mohanty*, Perez~Liebana*, Salakhutdinov*, Topin*, Veloso*, and
  Wang*]{gussminerlneurips2019}
William~H. Guss, Cayden Codel*, Katja Hofmann*, Brandon Houghton*, Noboru
  Kuno*, Stephanie Milani*, Sharada Mohanty*, Diego Perez~Liebana*, Ruslan
  Salakhutdinov*, Nicholay Topin*, Manuela Veloso*, and Phillip Wang*.
\newblock The {M}ine{RL} competition on sample efficient reinforcement learning
  using human priors.
\newblock In \emph{The 33rd Conference on Neural Information Processing Systems
  Competition Track}, 2019.

\bibitem[Guss* et~al.(2019)Guss*, Houghton*, Topin, Wang, Codel, Veloso, and
  Salakhutdinov]{gussminerlijcai2019}
William~H. Guss*, Brandon Houghton*, Nicholay Topin, Phillip Wang, Cayden
  Codel, Manuela Veloso, and Ruslan Salakhutdinov.
\newblock Mine{RL}: A large-scale dataset of {M}inecraft demonstrations.
\newblock In \emph{The 28th International Joint Conference on Artificial
  Intelligence}, 2019.

\bibitem[Hessel et~al.(2018)Hessel, Modayil, Van~Hasselt, Schaul, Ostrovski,
  Dabney, Horgan, Piot, Azar, and Silver]{hessel2018rainbow}
Matteo Hessel, Joseph Modayil, Hado Van~Hasselt, Tom Schaul, Georg Ostrovski,
  Will Dabney, Dan Horgan, Bilal Piot, Mohammad Azar, and David Silver.
\newblock Rainbow: Combining improvements in deep reinforcement learning.
\newblock In \emph{The 32nd AAAI Conference on Artificial Intelligence}, 2018.

\bibitem[Hester et~al.(2018)Hester, Vecerik, Pietquin, Lanctot, Schaul, Piot,
  Horgan, Quan, Sendonaris, Osband, et~al.]{hester2018deep}
Todd Hester, Matej Vecerik, Olivier Pietquin, Marc Lanctot, Tom Schaul, Bilal
  Piot, Dan Horgan, John Quan, Andrew Sendonaris, Ian Osband, et~al.
\newblock Deep q-learning from demonstrations.
\newblock In \emph{The 32nd AAAI Conference on Artificial Intelligence}, 2018.

\bibitem[Ho and Ermon(2016)]{gail_2016}
Jonathan Ho and Stefano Ermon.
\newblock Generative adversarial imitation learning.
\newblock In \emph{Advancements in Neural Information Processing Systems},
  2016.

\bibitem[Houghton et~al.(2020)Houghton, Milani, Topin, Guss, Hofmann,
  Perez-Liebana, Veloso, and Salakhutdinov]{houghton2020guaranteeing}
Brandon Houghton, Stephanie Milani, Nicholay Topin, William Guss, Katja
  Hofmann, Diego Perez-Liebana, Manuela Veloso, and Ruslan Salakhutdinov.
\newblock Guaranteeing reproducibility in deep learning competitions.
\newblock In \emph{The 23rd Conference on Neural Information Processing
  Systems, Challenges in Machine Learning (CiML) Workshop}, 2020.

\bibitem[Hsu(2019)]{hsu_2019}
Jeremy Hsu.
\newblock Ai takes on popular minecraft game in machine-learning contest, Nov
  2019.
\newblock URL \url{https://www.nature.com/articles/d41586-019-03630-0}.

\bibitem[Johnson et~al.(2016)Johnson, Hofmann, Hutton, and
  Bignell]{johnson2016malmo}
Matthew Johnson, Katja Hofmann, Tim Hutton, and David Bignell.
\newblock The malmo platform for artificial intelligence experimentation.
\newblock In \emph{The 25th International Joint Conference on Artificial
  Intelligence}, pages 4246--4247, 2016.

\bibitem[Kidzi{\'n}ski et~al.(2018)Kidzi{\'n}ski, Mohanty, Ong, Hicks, Carroll,
  Levine, Salath{\'e}, and Delp]{kidzinski2018learning}
{\L}ukasz Kidzi{\'n}ski, Sharada~P Mohanty, Carmichael~F Ong, Jennifer~L Hicks,
  Sean~F Carroll, Sergey Levine, Marcel Salath{\'e}, and Scott~L Delp.
\newblock Learning to run challenge: Synthesizing physiologically accurate
  motion using deep reinforcement learning.
\newblock In \emph{The NIPS'17 Competition: Building Intelligent Systems},
  pages 101--120. Springer, 2018.

\bibitem[Milani et~al.(2020)Milani, Topin, Houghton, Guss, Mohanty, Nakata,
  Vinyals, and Kuno]{milani2020minerl}
Stephanie Milani, Nicholay Topin, Brandon Houghton, William~H. Guss, Sharada~P.
  Mohanty, Keisuke Nakata, Oriol Vinyals, and Noboru~Sean Kuno.
\newblock Retrospective analysis of the 2019 {MineRL} competition on sample
  efficient reinforcement learning.
\newblock \emph{Proceedings of Machine Learning Research: NeurIPS 2019
  Competition and Demonstration Track}, 2020.

\bibitem[Mnih et~al.(2015)Mnih, Kavukcuoglu, Silver, Rusu, Veness, Bellemare,
  Graves, Riedmiller, Fidjeland, Ostrovski, et~al.]{mnih2015human}
Volodymyr Mnih, Koray Kavukcuoglu, David Silver, Andrei~A Rusu, Joel Veness,
  Marc~G Bellemare, Alex Graves, Martin Riedmiller, Andreas~K Fidjeland, Georg
  Ostrovski, et~al.
\newblock Human-level control through deep reinforcement learning.
\newblock \emph{Nature}, 518\penalty0 (7540):\penalty0 529, 2015.

\bibitem[Mnih et~al.(2016)Mnih, Badia, Mirza, Graves, Lillicrap, Harley,
  Silver, and Kavukcuoglu]{mnih2016asynchronous}
Volodymyr Mnih, Adria~Puigdomenech Badia, Mehdi Mirza, Alex Graves, Timothy
  Lillicrap, Tim Harley, David Silver, and Koray Kavukcuoglu.
\newblock Asynchronous methods for deep reinforcement learning.
\newblock In \emph{The 33rd International Conference on Machine Learning},
  pages 1928--1937, 2016.

\bibitem[Nichol et~al.(2018)Nichol, Pfau, Hesse, Klimov, and
  Schulman]{nichol2018gotta}
Alex Nichol, Vicki Pfau, Christopher Hesse, Oleg Klimov, and John Schulman.
\newblock Gotta learn fast: A new benchmark for generalization in rl.
\newblock \emph{arXiv preprint arXiv:1804.03720}, 2018.

\bibitem[Oh et~al.(2016)Oh, Chockalingam, Singh, and Lee]{oh2016control}
Junhyuk Oh, Valliappa Chockalingam, Satinder Singh, and Honglak Lee.
\newblock Control of memory, active perception, and action in {M}inecraft.
\newblock \emph{arXiv preprint arXiv:1605.09128}, 2016.

\bibitem[OpenAI(2018)]{openai_2018}
OpenAI.
\newblock Openai five, Sep 2018.
\newblock URL \url{https://blog.openai.com/openai-five/}.

\bibitem[Panse et~al.(2018)Panse, Madheshia, Sriraman, and
  Karande]{panse2018imitation}
Ameya Panse, Tushar Madheshia, Anand Sriraman, and Shirish Karande.
\newblock Imitation learning on atari using non-expert human annotations.
\newblock 2018.

\bibitem[Paszke et~al.(2019)Paszke, Gross, Massa, Lerer, Bradbury, Chanan,
  Killeen, Lin, Gimelshein, Antiga, Desmaison, Kopf, Yang, DeVito, Raison,
  Tejani, Chilamkurthy, Steiner, Fang, Bai, and Chintala]{pytorch}
Adam Paszke, Sam Gross, Francisco Massa, Adam Lerer, James Bradbury, Gregory
  Chanan, Trevor Killeen, Zeming Lin, Natalia Gimelshein, Luca Antiga, Alban
  Desmaison, Andreas Kopf, Edward Yang, Zachary DeVito, Martin Raison, Alykhan
  Tejani, Sasank Chilamkurthy, Benoit Steiner, Lu~Fang, Junjie Bai, and Soumith
  Chintala.
\newblock Pytorch: An imperative style, high-performance deep learning library.
\newblock In \emph{Advances in Neural Information Processing Systems 32}, pages
  8024--8035. 2019.

\bibitem[Perez-Liebana et~al.(2019)Perez-Liebana, Hofmann, Mohanty, Kuno,
  Kramer, Devlin, Gaina, and Ionita]{perez2019multi}
Diego Perez-Liebana, Katja Hofmann, Sharada~Prasanna Mohanty, Noburu Kuno,
  Andre Kramer, Sam Devlin, Raluca~D Gaina, and Daniel Ionita.
\newblock {The Multi-Agent Reinforcement Learning in Malm\"{O} (MARL\"{O})
  Competition}.
\newblock \emph{arXiv preprint arXiv:1901.08129}, 2019.

\bibitem[Reddy et~al.(2019)Reddy, Dragan, and Levine]{reddy2019sqil}
Siddharth Reddy, Anca~D Dragan, and Sergey Levine.
\newblock Sqil: Imitation learning via reinforcement learning with sparse
  rewards.
\newblock \emph{arXiv preprint arXiv:1905.11108}, 2019.

\bibitem[Salge et~al.(2018)Salge, Green, Canaan, and
  Togelius]{salge2018generative}
Christoph Salge, Michael~Cerny Green, Rodgrigo Canaan, and Julian Togelius.
\newblock {Generative Design in Minecraft (GDMC): Settlement Generation
  Competition}.
\newblock In \emph{The 13th International Conference on the Foundations of
  Digital Games}, page~49. ACM, 2018.

\bibitem[Schaul et~al.(2015)Schaul, Quan, Antonoglou, and
  Silver]{schaul2015prioritized}
Tom Schaul, John Quan, Ioannis Antonoglou, and David Silver.
\newblock Prioritized experience replay.
\newblock In \emph{Proceedings of the International Conference on Learning
  Representations}, 2015.

\bibitem[Schulman et~al.(2017)Schulman, Wolski, Dhariwal, Radford, and
  Klimov]{ppo}
John Schulman, Filip Wolski, Prafulla Dhariwal, Alec Radford, and Oleg Klimov.
\newblock Proximal policy optimization algorithms.
\newblock \emph{arXiv preprint arXiv:1707.06347}, 2017.

\bibitem[Shead(2019)]{shead_2019}
Sam Shead.
\newblock Minecraft diamond challenge leaves ai creators stumped, Dec 2019.
\newblock URL \url{https://www.bbc.com/news/technology-50720823}.

\bibitem[Shu et~al.(2017)Shu, Xiong, and Socher]{shu2017hierarchical}
Tianmin Shu, Caiming Xiong, and Richard Socher.
\newblock Hierarchical and interpretable skill acquisition in multi-task
  reinforcement learning.
\newblock \emph{arXiv preprint arXiv:1712.07294}, 2017.

\bibitem[Silver et~al.(2017)Silver, Schrittwieser, Simonyan, Antonoglou, Huang,
  Guez, Hubert, Baker, Lai, Bolton, et~al.]{silver2017mastering}
David Silver, Julian Schrittwieser, Karen Simonyan, Ioannis Antonoglou, Aja
  Huang, Arthur Guez, Thomas Hubert, Lucas Baker, Matthew Lai, Adrian Bolton,
  et~al.
\newblock Mastering the game of go without human knowledge.
\newblock \emph{Nature}, 550\penalty0 (7676):\penalty0 354--359, 2017.

\bibitem[Silver et~al.(2018)Silver, Hubert, Schrittwieser, Antonoglou, Lai,
  Guez, Lanctot, Sifre, Kumaran, Graepel, et~al.]{alphazero}
David Silver, Thomas Hubert, Julian Schrittwieser, Ioannis Antonoglou, Matthew
  Lai, Arthur Guez, Marc Lanctot, Laurent Sifre, Dharshan Kumaran, Thore
  Graepel, et~al.
\newblock A general reinforcement learning algorithm that masters chess, shogi,
  and go through self-play.
\newblock \emph{Science}, 362, 2018.

\bibitem[Synced(2019)]{synced_2019}
Synced.
\newblock Neurips 2019 will host minecraft reinforcement learning competition,
  May 2019.
\newblock URL
  \url{https://medium.com/syncedreview/neurips-2019-will-host-minecraft-reinforcement-learning-competition-146e8bc8da1}.

\bibitem[Tessler et~al.(2017)Tessler, Givony, Zahavy, Mankowitz, and
  Mannor]{tessler2017deep}
Chen Tessler, Shahar Givony, Tom Zahavy, Daniel~J Mankowitz, and Shie Mannor.
\newblock A deep hierarchical approach to lifelong learning in minecraft.
\newblock In \emph{The 31st AAAI Conference on Artificial Intelligence}, 2017.

\bibitem[van Hasselt et~al.(2016)van Hasselt, Guez, and
  Silver]{hasselt2015deep}
Hado van Hasselt, Arthur Guez, and David Silver.
\newblock Deep reinforcement learning with double q-learning.
\newblock In \emph{The 30th AAAI Conference on Artificial Intelligence}, 2016.

\bibitem[Van~Hasselt et~al.(2016)Van~Hasselt, Guez, and Silver]{van2016deep}
Hado Van~Hasselt, Arthur Guez, and David Silver.
\newblock Deep reinforcement learning with double q-learning.
\newblock 2016.

\bibitem[Vincent(2019)]{vincent_2019}
James Vincent.
\newblock Ai has bested chess and go, but it struggles to find a diamond in
  minecraft, Dec 2019.
\newblock URL
  \url{https://www.theverge.com/2019/12/13/21020230/ai-minecraft-minerl-diamond-challenge-microsoft-reinforcement-learning}.

\bibitem[Vinyals et~al.(2019)Vinyals, Babuschkin, Czarnecki, Mathieu, Dudzik,
  Chung, Choi, Powell, Ewalds, Georgiev, et~al.]{starcraft2019}
Oriol Vinyals, Igor Babuschkin, Wojciech~M. Czarnecki, Michael Mathieu, Andrew
  Dudzik, Junyong Chung, David~H. Choi, Richard Powell, Timo Ewalds, Petko
  Georgiev, et~al.
\newblock Grandmaster level in {StarCraft II} using multi-agent reinforcement
  learning.
\newblock \emph{Nature}, 2019.

\bibitem[Wang et~al.(2016)Wang, Schaul, Hessel, Hasselt, Lanctot, and
  Freitas]{wang2016dueling}
Ziyu Wang, Tom Schaul, Matteo Hessel, Hado Hasselt, Marc Lanctot, and Nando
  Freitas.
\newblock Dueling network architectures for deep reinforcement learning.
\newblock In \emph{Proceedings of the International Conference on Machine
  Learning}, 2016.

\end{thebibliography}

\end{document}